\documentclass[12pt]{article}
\usepackage{geometry}
\usepackage{microtype}
\usepackage{graphicx}
\usepackage{xcolor}
\usepackage{subcaption}
\usepackage{booktabs} 
\usepackage{stmaryrd}
\usepackage[american]{babel}
\usepackage{hyperref}
\usepackage[ruled,vlined]{algorithm2e}
\usepackage{tabularx}
\usepackage{array}
\usepackage{authblk}
\usepackage{natbib}
\usepackage{tikz}
\usetikzlibrary{arrows.meta,positioning,calc,decorations.pathmorphing,fit,backgrounds}
\usepackage{amsmath}
\usepackage{amssymb}
\usepackage{mathtools}
\usepackage{amsthm}
\usepackage{bbm}
\usepackage{standalone}
\usepackage{enumitem}

\theoremstyle{plain}
\newtheorem{theorem}{Theorem}[section]

\newtheorem{lemma}[theorem]{Lemma}

\theoremstyle{definition}

\theoremstyle{remark}

\renewcommand{\P}{\mathbb{P}}
\newcommand{\E}[2][]{\mathbb{E}_{#1} \!\left[#2\right]}
\newcommand{\abs}[1]{\left|#1\right|}
\newcommand{\F}{\mathcal{F}}
\renewcommand{\H}{\mathcal{H}}

\newcommand{\KL}[2][]{\mathbb{KL}_{#1}\!\left[#2\right]}

\newcolumntype{Y}{>{\centering\arraybackslash}X}

\DeclareMathOperator*{\argmax}{arg\,max}

\graphicspath{{./}}

\newcommand{\xinlong}[1]{}
\newcommand{\harsha}[1]{}
\newcommand{\vinayak}[1]{}

\newcommand{\mugt}{\mu_{\text{gt}}}

\title{Neural Diffusion Intensity Models for Point Process Data}

\author[1]{Xinlong Du}
\author[1]{Harsha Honnappa}
\author[2]{Vinayak Rao}
\affil[1]{%
    Edwardson School of Industrial Engineering, Purdue University
}
\affil[2]{%
    Department of Statistics, Purdue University
}

\date{}

\begin{document}
\maketitle
\begin{abstract}

Cox processes model overdispersed point process data via a latent stochastic intensity, but both nonparametric estimation of the intensity model and posterior inference over intensity paths are typically intractable, relying on expensive MCMC methods. We introduce Neural Diffusion Intensity Models, a variational framework for Cox processes driven by neural SDEs. Our key theoretical result, based on enlargement of filtrations, shows that conditioning on point process observations preserves the diffusion structure of the latent intensity with an explicit drift correction. This guarantees the variational family contains the true posterior, so that ELBO maximization coincides with maximum likelihood estimation under sufficient model capacity. We design an amortized encoder architecture that maps variable-length event sequences to posterior intensity paths by simulating the drift-corrected SDE, replacing repeated MCMC runs with a single forward pass. Experiments on synthetic and real-world data demonstrate accurate recovery of latent intensity dynamics and posterior paths, with orders-of-magnitude speedups over MCMC-based methods.
\end{abstract}

\section{Introduction}

\begin{figure}[t]
    \centering    \includegraphics[width=0.7\linewidth]{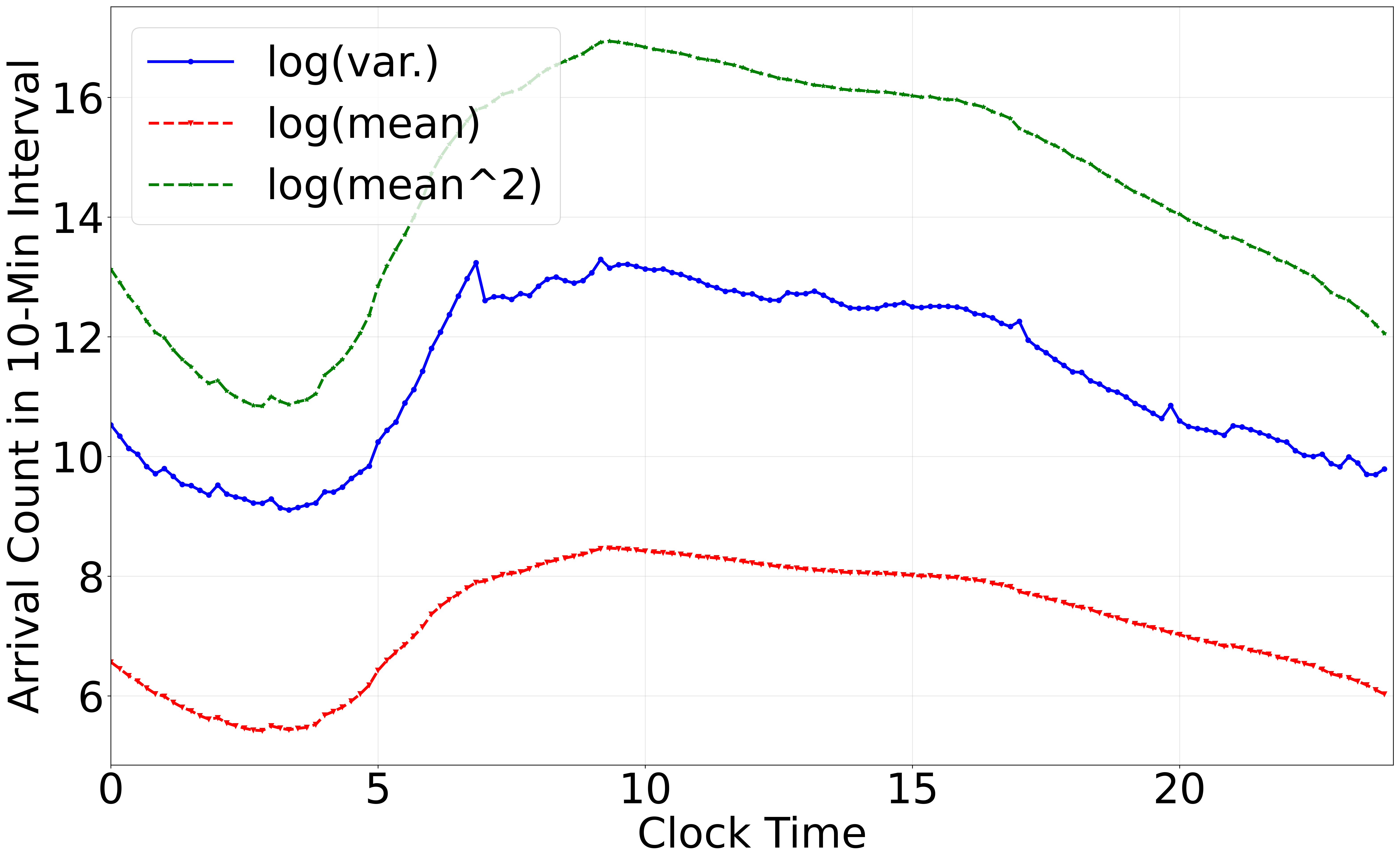}
    \caption{The variance curve exceeds the mean curve, indicating strong overdispersion of the arrival point process, in the US Bank data from~\citep{seelab2024data}.}
    \label{fig:placeholder}\label{fig:overdisperse-stats}
\end{figure}

Point process data appear in many scientific and industrial applications, including neural spike trains~\citep{amemori2001gaussian}, social interactions~\citep{PerryWolfe2013}, finance and insurance models~\citep{Bjork2021} and queuing models~\citep{Bremaud1981}. Such data are sparse, irregular, and often exhibit substantial \emph{over-dispersion} relative to simple inhomogeneous Poisson models \citep{Jongbloed2001, Avramidis2005}. 
As an example, in Section~\ref{sec:us_bank}, we will consider a dataset of call records from a large U.S. bank call center.  Figure~\ref{fig:overdisperse-stats} shows  the  statistics of the number of arrivals every 10 minutes: this exhibits a very obvious overdispersion.
A natural and expressive framework is to model these temporal events as Cox processes, wherein event arrivals follow a Poisson process with a latent \emph{stochastic} intensity~\citep[Chapter 6]{daley2003pointprocessesI}. {The randomness of the intensity function helps explain the greater between-realization variability that is characteristic of overdispersed  point processes. 

In this paper, we model the intensity process as Markov diffusions that are solutions to unknown stochastic differential equations (SDEs).} In this case the point process model is commonly referred to as a diffusion-driven Cox process (DDCP)~\citep{Goncalves2024}. 
The Markovianity of the intensity process is a common requirement in many real-world applications; for instance,~\citep{zhang2014scaling} propose a Cox-Ingersoll-Ross (CIR) diffusion model of the intensity of the U.S. bank data in Figure~\ref{fig:overdisperse-stats}.
{Modeling via an SDE allows one to explicitly encode dynamic mechanisms, 
such as mean reversion or state-dependent volatility. These provide an interpretable description of how the event rate evolves over multiple time-scales, making them particularly appealing when the intensity is believed to follow an underlying economic, physical, or biological process.}

Learning a DDCP amounts to estimating the coefficients of the latent SDE from point process observations. This is a challenging problem, 
involving intractable infinite-dimensional integrals over paths and requiring 
computationally intensive Markov chain Monte Carlo (MCMC) methods~\citep{Goncalves2024}. 
These challenges persist even after fitting the model: now, computing the posterior of the intensity conditioned on a point process observation requires expensive MCMC simulations. 
In this work, we introduce \textbf{Neural Diffusion Intensity Models}, a nonparametric variational framework for Cox processes driven by latent SDE intensity processes. Our approach combines three key ideas:

\begin{itemize}[leftmargin=1em]
\item[1.] \textbf{Neural SDE priors.}
We parametrize the drift of the latent SDE with a neural network, yielding a flexible generative model for intensity paths. We do not emphasize this here, but our methodology applies equally in settings with standard parametric SDE drift functions, or semiparametric models that sit in between.
\item[2.] \textbf{Posterior characterization.} 
Using tools from \textit{enlargement of filtrations} (EoF), we show that conditioning on point process observations introduces a score-like {\it drift correction} while preserving the diffusion structure of the latent intensity. That is, the posterior intensity process remains a diffusion with the same diffusion coefficient and a modified score-type drift. This establishes a conjugacy at the level of path measures. 
\item[3.] \textbf{Path-space variational inference.}
Motivated by the structural result above, we construct a variational family {indexed by the above drift correction, each index corresponding to a different approximation to the posterior intensity process. We parametrize the drift correction as a neural network, so that for expressive enough neural network architectures,} the variational family contains the true posterior. As a consequence, maximizing the evidence lower bound (ELBO) coincides with maximum likelihood estimation. 
\end{itemize}

Viewed as a VAE, the drift neural network is a decoder from Brownian paths to point process observations via the intermediate intensity, while the drift correction neural network forms an encoder mapping from point process observations to  posterior distributions over intensity paths. Under our framework, posterior inference is amortized: given point process observations, posterior intensity paths can be sampled by simulating the drift-corrected SDE directly.

\subsection{Related Work}

{There have been numerous prior works on point process models with stochastic intensity functions. A large number of these model the intensity function as a transformed realization of a smooth Gaussian process prior, and carry out posterior inference via MCMC~\citep{moller1998log, adams2009tractable}, variational inference~\citep{donner2018efficient, aglietti2019structured}, or approaches like INLA~\citep{simpson2016going}. Here, the latent intensity primarily serves as a smoothing mechanism, 
and these Gaussian process models lack the capacity to represent mechanistic dynamics in the way that a drift term in an SDE can.

Closer to our work is~\citep{duncker2019learning,jaiswal2021,wang2020estimating}, all of whom consider DDCPs, and attempt to flexibly learn the drift-function of the underlying SDE using variational Bayes. In the former, the authors make a Gaussian approximation to the posterior along the lines of~\citep{archambeau2007gaussian} (see below): however this 
 introduces a variational gap. 
 \citep{jaiswal2021,wang2020estimating} proposed a variational inference scheme for DDCPs, using a stochastic partial differential equation satisfied by the smoothing posterior density as a heuristic to identify the posterior diffusion. The present paper supersedes this work by providing a rigorous EoF-based structural characterization of the posterior path measure (not considered in~\citep{jaiswal2021,wang2020estimating}), and establishes that a sufficiently expressive variational family will contain the true posterior.~\citep{wang2021calibrating} extended the method in~\citep{wang2020estimating} to a Poisson random measure likelihood, slightly generalizing the setting in~\citep{jaiswal2021,wang2020estimating}.}
Furthermore, these works do not consider amortized inference, necessitating a fresh inference run for each new observation.

A different approach is that taken by neural temporal point processes (neural TPPs) and related models~\citep{du2016recurrent,shchur2021neural,zuo2020transformer,mei2017neural,shchur2020fast}: these model marked point processes autoregressively, predicting the time and mark of the next event in a sequence 
using neural network architectures like RNNs, GRUs or LSTMs. 
Works like APP-VAE \citep{mehrasa2019variational}, VEPP \citep{pan2020variational} and \citep{shibue2024marked} can be viewed as latent variable neural TPPs, and apply variational autoencoder ideas to point process data by introducing latent variables at the level of inter-arrival times.
These models, while flexible, are geared more towards prediction. They are non-intensity-based models, unable to model, learn the dynamics of, or capture posterior uncertainty about the latent stochastic intensity process. Even here, the ones that utilize variational inference/autoencoders do not emphasize the structure of the `true' posterior, and since their variational families are Gaussian, 
a variational gap is present structurally. 

{
Moving our focus from point processes to SDEs,  there is an extensive literature on MCMC~\citep{beskos2006exact,Goncalves2024} and variational inference for continuous-time stochastic models. 
The latter can be traced to the seminal work in~\citep{archambeau2007gaussian}, who first proposed approximating the posterior of a diffusion process with a linear SDE, deriving a variational lower bound via Girsanov's theorem.
Works like \citep{Opper2019} and \citep{Ryder2018} build on these ideas 
for SDE latent paths, and share some ideas with our work. For example,  ~\citep{Opper2019} assumes (like us) that the approximate posterior path is an SDE with a learnable drift, and proposes to learn the posterior drift as a stochastic optimal control problem. \citep{Opper2019} however does not consider point processes observations (which present special challenges), does not attempt to identify the `true' posterior, nor is the question of amortized inference considered.
 Latent Neural SDEs for generative modeling have also been studied in~\citep{kidger2021neural,li2020scalable}, however none of these works consider point process observations generated by the latent-diffusion.
}


Opper's formulation of variational inference as stochastic optimal control \citep{Opper2019} is closely related to \citep{Tzen2019}, who derive the {Föllmer} drift — the drift that exactly samples from the posterior over the terminal state 
$X_T$ given observations $Y \sim P(\cdot|X_T)$ — as the solution to an optimal control problem, parameterized by a neural network. This is free of variational gap at the level of the terminal law. In our setting, however, the point process observations depend on the full intensity trajectory, so we must match the entire path measure rather than just the terminal law. Our EoF-based result provides the path-measure analogue of their terminal-law result. However, observe that we identify the posterior drift correction from purely probabilistic considerations rather than as an optimal control. 


{The question of amortized inference in dynamical latent variable models has been studied in the discrete-time setting by \citep{krishnan2017structured} and \citep{marino2018general}, who show that conditioning the approximate posterior on the full observation sequence can yield substantially better posterior approximations than causal (filtering) conditioning. 
Our work can be viewed as the continuous-time, point-process analogue of this principle, with the additional theoretical results from EoF about the structural necessity of  full-sequence conditioning. 
}
\section{Problem Formulation}
\vspace{0em}

{We formulate our model on a filtered probability space $(\Omega, \mathcal{F}, 
(\mathcal{F}_t)_{t \in [0,T]}, \mathbb{P}^\star)$, where $\mathbb{P}^\star$ 
denotes the unknown ground-truth data-generating measure. We observe event data 
as a simple point process~\citep[Chapter 1]{daley2003pointprocessesI} $X = \{\tau_i\}_{i=1}^K$ on $[0,T]$, with $0 < 
\tau_1 < \cdots < \tau_K \leq T$ and $K$ a random variable giving the number 
of events. We may use $X$ interchangeably with the counting process $N_{0:T}$ where $N_t:=\sum_i \mathbf{1}_{\{\tau_i\le t\}}$. We let $\mu_{\mathrm{gt}} = \mathbb{P}^\star \circ X^{-1}$ denote 
the ground-truth distribution over event sequences, and let $\mathcal X := \cup_{k \in \mathbb N} \mathbb R^{k}$ represent the sample space for $X$.}
{The filtration $(\mathcal{F}_t)$ 
is assumed to carry both the observed events $X$ and a latent intensity process 
$Z = (Z_t)_{t \in [0,T]}$, which is $(\mathcal{F}_t)$-adapted 
but not directly observed. Let $\mathcal Z := C([0,T];\mathbb R^d)$ represent the sample space of $\mathbb R^d$-valued continuous functions for $Z$.

\subsection{Neural Diffusion Intensity Model}\label{sec:neural-cox}

We focus on one-dimensional simple point processes, noting that our framework 
can be extended to multivariate point processes
by replacing the scalar 
intensity process with a vector-valued SDE\footnote{This would model 
of multiple event streams that may share a common latent intensity 
or evolve independently.} 
 on $[0,T]$. Our model class $\{P_\theta, \theta \in \Theta\}$ consists of Cox processes (doubly stochastic Poisson processes), defined by the hierarchical structure
$P_\theta(X) := \int_{\mathcal Z} dP_\theta(X,Z)$ where $P_\theta(X,Z)=P(X|Z)P_\theta(Z)$, $P(X|Z)$ and $P_\theta(Z)$ are path measures and $\Theta \subseteq \mathbb R^p,~p > 0$.  
%
Conditioned on a latent intensity trajectory $Z = (Z_t)_{t\in[0,T]}$, the observation $X$ is an inhomogeneous Poisson process with rate  $(Z_t)_{t\in[0,T]}$. The prior measure $P_\theta(Z)$ over intensities is induced by the Markov diffusion solution of the SDE
\begin{equation}
\label{eq:sde}
dZ_t = b_\theta(Z_t,t)\,dt + \sigma(Z_t,t)\,dB_t,
\qquad Z_0 = z.
\end{equation}
Here $B_t$ is a Brownian motion adapted to $(\F_t)_{t\in [0,T]}$, while
\begin{itemize}
    \item $b_\theta$ is a parameterized drift function (modeled by a neural network),
    \item $\sigma$
    is a known diffusion coefficient (e.g., fixed as a constant or set by the CIR model).
\end{itemize}
We assume that the drift and diffusion coefficients satisfy the usual conditions consistent for a strong solution of the SDE to exist; see~\citep[Theorem~5.2.1]{oksendal2003stochastic}. Then, the joint measure $P_\theta(X, Z)$ is well-defined on $\mathcal{X} \times \mathcal{Z}$ provided $Z_t \geq 0$ a.s., making it is a valid intensity function. 
This construction yields an infinite mixture model~\citep{Lindsay1995} of Poisson processes~\citep{moller1998log}, with mixing distribution given by the SDE prior $P_\theta(Z)$. See Section~\ref{sec:neural-cox-deets} for the measure-theoretic desiderata for the model.
\subsection{Nonparametric Maximum Likelihood}

With this setup, learning $P_\theta(X)$ reduces to estimating the drift parameter $\theta$.}
We do this by maximizing the expected log-likelihood under the ground truth distribution $\mugt$: 
$$\sup_{\theta \in \Theta}\E[\mu_{\text{gt}}]{\log P_\theta(X)} = \sup_{\theta \in \Theta}\E[\mu_{\text{gt}}]{\log \int P(X|Z) P_\theta(dZ)}.$$
This is a nonparametric maximum likelihood estimation (NPMLE) problem over an infinite mixture model; see \citep[Chapter 5]{Lindsay1995}. 
{We assume that the model class $\{P_\theta, \theta \in \Theta\}$ is 
\emph{well-specified} in the sense that $\mathrm{KL}(\mu_{\text{gt}} \| 
P_\theta) < \infty$ for some $\theta \in \Theta$. This ensures that $\mathbb{E}_{\mu_{\text{gt}}}[\log P_\theta(X)]$ 
is well-defined and finite, and that the maximum likelihood objective 
is not trivially $-\infty$.}

In practice, $\mugt$ is unknown and we assume access to $n$ i.i.d. observations of the data, i.e., $\{X^{i}\}_{i=1}^n$, with $X^i\overset{\text{i.i.d.}}{\sim}\mugt$, leading to the empirical objective
\begin{align}
\sup_{\theta \in \Theta} \E[\hat\mu]{\log \int P(X|Z) dP_\theta(Z)} \label{eq:emp_obj}
\end{align}
where $\hat{\mu}:=\frac{1}{n}\sum_{i=1}^n\delta_{X^{i}}$ is the empirical measure. 

Solving this maximization problem however involves a structural difficulty: with a few special exceptions, the integral on the RHS is intractable. This 
has been classically addressed in the mixture modeling literature by resorting to an expectation-maximization (EM) type algorithm; see Appendix~\ref{sec:em} and {~\citep[Chapter 3 and 6]{Lindsay1995} for finite dimensional settings. 

In our setting, the E-step requires sampling from the posterior path measure $P_\theta(Z|X)$, typically via MCMC, while the M-step updates the prior measure $\theta$. This can be computationally expensive, since every gradient step requires new MCMC samples from the path posterior distribution. Furthermore,
the EM algorithm will only learn the prior dynamics, and posterior inference for a newly observed point process will require additional MCMC at test time. Since the posterior is a path space measure, such MCMC sampling can be challenging, especially in settings where repeated or real-time inference is needed. 

To bypass this computational burden, 
we introduce an amortized approximation to the posterior. We model the posterior intensity process as the solution of the SDE
\begin{equation}\label{eq:posterior_sde}
\begin{aligned}
dZ_t&=\Big[b_\theta(Z_t, t)+\mathbf{1}_{\{t\le T'\}}\sigma(Z_t,t)u_\beta\left(Z_t, t,T', N_{0:T'}\right)\Big]dt\\
&\quad+\sigma(Z_t,t)\,dB_t.
\end{aligned}
\end{equation}
and treat the path measure induced by this SDE as the variational family $Q_\phi(Z|X)$, where $\phi=(\theta,\beta)$. 
This construction preserves the diffusion coefficient, while introducing an observation-dependent drift correction through $u_\beta$. Once $b_\theta$ and $u_\beta$ are learned, posterior inference for any observation reduces to simulating sample paths from Eq.~\eqref{eq:posterior_sde}. Posterior inference becomes fully amortized. 


\subsection{Enlargement of Filtrations}~\label{sec:eof-main}
At this point, this structure imposed by our variational approximation remains a heuristic. There are two fundamental questions to consider regarding the true posterior process:

\begin{itemize}
	\item[i.] (\textbf{Existence}) Does the posterior intensity, conditioned on a point process observation, remain a diffusion with an SDE representation? And if so:
	\item[ii.] (\textbf{Structure}) How does its drift relate to the prior drift? 
\end{itemize}

Answering the first question justifies the validity of using path measures induced by Markov diffusions (as solutions of SDEs) for the approximate posteriors, and answering the second sheds light on the correct functional form of the posterior drift correction. 

Both questions can be answered using tools from {Enlargement of Filtrations} (EoF) ~\citep{Jeanblanc2009,Jacod2006}, wherein one studies how semimartingale decompositions change when the filtration is enlarged to include additional information, typically information about future events. In our setting, the enlargement corresponds to conditioning on the observed point process trajectory $X=N_{0:T'}$, where $T' \leq T$. Formally, we pass from the original filtration $(\F_t)$ to the ``enlarged'' filtration
$\mathcal{G}_t=\F_t\lor\sigma(N_{0:T'}),$
so that the future event times up to $T'$ are known at all earlier times. This is the continuous-time analogue of \emph{smoothing} in a hidden 
Markov model (HMM), where the latent state at time $t$ is inferred given the entire observation sequence rather than only the observations up to time $t$; here, the role of the full observation sequence is played by the complete point process trajectory $N_{0:T'}$.

Under Jacod's absolute continuity condition (Theorem~\ref{thm:jacod_condition}), one can apply the general semimartingale decomposition theorem (Theorem.~\ref{thm:decomposition}) to obtain the following result\footnote{Proof in Appendix \ref{sec:intense-enlarge}}:
\begin{theorem}\label{thm:jacod}
Fix $T' \leq T$ and let $N_{0:T'}$ be a one-dimensional point process on the interval $[0,T']$, and suppose the prior intensity process satisfies
$$dZ_t=b(Z_t,t)\,dt+\sigma(Z_t,t)\,dB_t, \quad Z_0=z_0.$$
Then, conditioned on the observed event times $X=(0<\tau_1<\cdots<\tau_{N_{T'}}\le T)$, the posterior intensity process admits the SDE representation 
\begin{align*}
dZ_t&=\big[b(Z_t,t)+\mathbf{1}_{\{t\le T'\}}\sigma(Z_t,t)^2h(Z_t,t,T',X)\big]dt\\
&\quad+\sigma(Z_t,t)\,d\tilde{B}_t, \quad Z_0=z_0,
\end{align*}
where $\tilde{B}$ is a Brownian motion with respect to 
$\mathcal{G}_t$ and 
\begin{equation}\label{eq:posterior_drift}
\begin{aligned}
&h(z, t, T', X):=\\
&\quad\quad\frac{\partial}{\partial z}\log\E{\exp\left(-\int_t^{T'}Z_s\,ds\right)\prod_{t<\tau_i\le T'}Z_{\tau_i}\Bigg|Z_t=z},
\end{aligned}
\end{equation}
with the conditional expectation with respect to the prior.
\end{theorem}

Theorem \ref{thm:jacod} answers the two structural questions: 

\begin{enumerate}
    \item[i.] \textbf{Diffusion structure is preserved.} 
    Under conditioning on the point process realization, the intensity remains a semi-martingale diffusion with the same diffusion coefficient. Only the drift is modified.
    \item[ii.] \textbf{Posterior correction is a score-type correction.}
    {The additional drift term $\sigma^2 h$ takes the form of a 
\emph{score function}: it is the gradient of the log-density of 
future observations with respect to the current state, evaluated 
under the prior. }
\end{enumerate}

{The posterior correction is analogous to the classifier guidance 
term in conditional diffusion models~\citep{dhariwal2021diffusion}, 
where the score of a classifier steers the prior toward the 
conditioned distribution.} Comparing Theorem~\ref{thm:jacod} with the variational SDE (Eq.~\eqref{eq:posterior_sde}) introduced earlier, we see that the amortized correction $u_\beta$ is intended to approximate the quantity
$\sigma(Z_t,t)h(Z_t,t,T',X)$.
The theorem therefore provides a principled, structured target for which variational inference methodology can be applied to learn. In this sense, EoF does not merely justify the SDE ansatz, but precisely identifies the functional object that amortization must approximate. {Under sufficient expressiveness of $u_\beta$, the variational family contains the true posterior, and the variational gap vanishes.}

\section{Variational Methodology}

Consider the empirical NPMLE objective~\eqref{eq:emp_obj}, 
with a single observation $X$, so that $\hat{\mu}=\delta_X$. For any probability measure $Q$ equivalent to $P_\theta$, i.e, $Q \sim P_\theta$, we write
$\log P_\theta(X)=\log\int P(X|Z)\frac{dP_\theta(Z)}{dQ(Z)}\,dQ(Z).$
Applying Jensen's inequality gives the standard evidence lower bound (ELBO)
\begin{align}
\log P_\theta(X)\ge \E[Q(Z)]{\log P(X|Z)}-\KL{Q(Z)||P_\theta(Z)}. \label{eq:elbo}
\end{align}
This bound is tight when $Q=P_\theta(Z|X)$. 
%
Recall the posterior drift correction $u_\beta(\cdot)$ indexed by $\beta$ from Eq.~\eqref{eq:posterior_sde}, and let $\phi := (\theta,\beta)$. Then ELBO takes 
the following specific form: 
\begin{theorem}\label{thm:elbo}
Suppose the prior model $\{P_\theta(Z)\}$ is specified by Eq.~\eqref{eq:sde}, and the variational family $\{Q_\phi\}$ is specified by Eq.~\eqref{eq:posterior_sde}, then the evidence lower bound for the log-likelihood reduces to $\mathcal{L}(\theta,\beta;X) = $
\begin{equation}\label{eq:elbo_vi}
\E[Q_\phi]{\log P(X|Z)-\frac{1}{2}\int_0^{T'}u_\beta^2(Z_t,t,T',X)\,dt}.
\end{equation}
\end{theorem}
We direct readers to Appendix.~\ref{sec:vi_elbo_derivation} for the proof of this.

To train our model, we jointly optimize the empirical ELBO,
\begin{equation}\label{eq:empirical_elbo}
    \frac{1}{n}\sum_{i=1}^n\mathcal{L}(\theta,\beta;X^i),
\end{equation}
with respect to $\theta$ and $\beta$, learning a generative prior over intensity trajectories and an amortized approximation to the  posterior SDE. Once trained, posterior inference for a new point pattern $X'$ reduces to simulating the SDE in Eq.~\eqref{eq:posterior_sde}, avoiding the need for repeated MCMC sampling.

\subsection{Optimization strategy}\label{sec:gradient_computations}

 To optimize the ELBO in Eq.~\eqref{eq:empirical_elbo}, we must first 
 evaluate the gradients  
$\partial_\theta\mathcal{L}(\theta,\beta;X)\text{ and }\partial_\beta\mathcal{L}(\theta,\beta;X)$ associated with any point process observation $X$ in Eq.~\eqref{eq:elbo_vi}.
Under standard regularity conditions, we can interchange differentiation and expectation to obtain \textbf{pathwise gradient representations:} the derivatives of $\mathcal{L}$ can be written as expectations under the variational path measure $Q_\phi(\cdot|X)$ of functionals involving the latent trajectory $Z$ and its \textbf{parameter sensitivities}, the Jacobian processes in~\eqref{eq:jac-theta}-\eqref{eq:jac-beta} $J^\theta_t:=(\partial_\theta Z)_t$ and $J^\beta_t:=(\partial_\beta Z)_t$ ({see Appendix.~\ref{sec:vi_gradients}}). More concretely, the likelihood part contributes to the gradients through gradients of the terms 
$\sum_{\tau_i\le T'}\log Z_{\tau_i} \text{ and } \int_0^{T'} Z_t\,dt$
as well as the KL term $\frac{1}{2}\int_0^{T'}u_\beta^2\,dt$. 
The gradients of the neural networks $b_\theta$ and $u_\beta$ with respect to the parameters and $Z$ can be computed using back-propagation (chain rule). 

At the implementation level, we numerically simulate the posterior SDE for $Z$ and the Jacobian processes using an Euler-Maruyama iteration on a grid $t_j=t\Delta t$, with common Brownian increments. 
 For each observation $X$, we simulate $m$ i.i.d. Brownian paths (discretized) and use these to simulate all the trajectories $(Z_{0:T'},J_{0:T'}^\theta, J_{0:T'}^\beta)$, and then average over these samples. Finally, we use stochastic gradient descent (SGD) on batched samples of the point process observations (specifically Adam). 

\subsection{Architectural Implications}

The prior drift $b_\theta$ is Markovian, therefore a standard neural network, such as a fully connected multi-layer perceptron (MLP), suffices for parametrization. The posterior correction, however, is structurally different. From Theorem \ref{thm:jacod}, the true drift correction $h$ depends on 1) the current state $Z_t$, 2) the current time $t$, 3) the terminal horizon $T'$, and 4) the set of future event times $\{\tau_i:t<\tau_i\le T'\}$. 
 Consequently, $u_\beta$ must take as input a varying-sized collection of event times $X$. While we could map $X$ to a fixed-dimensional summary statistic (such as the number of arrivals in $(t,T']$), this will necessarily discard the temporal structure of the data, and lead to an amortization gap.
To preserve all the information carried by a point process observation, we adapt the Deep Sets architecture \citep{Zaheer2018}, 
and parametrize the posterior correction with the following two-stage `pooling' architecture: 
\begin{align*}
&u_\beta(Z_t, t, T', N_{0:T'})=\\
&\quad\quad\quad\sigma(Z_t,t)\rho\left(t, T', \sum_{t<\tau_i\le T'}\psi(Z_t, \Delta_i, T'-\tau_i )\right),
\end{align*}
where $\Delta_i:=\tau_i-\tau_{i-1}$, and both $\rho$ and $\psi$ are fully connected MLPs. 

\section{Numerical Experiments}

In this section, we progressively evaluate our proposed variational inference method, including qualitative and quantitative analyses on synthetic datasets, as well as a real world dataset of call arrivals to a large U.S.-based bank call center. 
\subsection{Prior recovery}

We begin with a synthetic experiment in which the ground-truth intensity process is assumed to follow Cox–Ingersoll–Ross (CIR) dynamics. The CIR model is a natural choice because it ensures positivity of the intensity and is widely used in finance and operations research applications. Here, the latent intensity $Z_t$ evolves as
\begin{equation} dZ_t = 0.3(80 - Z_t)\,dt + \sqrt{Z_t}\,dB_t, \quad Z_0 = 5. \end{equation}

Using this model, we generate $256$ point process observations $\{X^i\}_{i=1}^{256}$ of the Cox process on the interval $[0,4]$ ({each coming from an independently sampled intensity path $Z$}). During training, 
The SDEs are simulated using the Euler-Maruyama scheme, with $100$ even steps. Monte Carlo gradient estimates are computed using { $10$ i.i.d. Brownian paths (discretized)} (details in appendix \ref{sec:vi_gradients}). The model is trained for $100$ epochs with a mini-batch size of $32$ and a learning rate of $0.005$ for both the prior drift network $b_\theta$ and the control network $u_\beta$. To stabilize optimization, gradients are clipped to have an $L_2$ norm of at most $5$. During training, the true prior dynamics are not accessible. 

Fig.~\ref{fig:prior_drift} compares the learned prior drift $b_\theta(Z_t,t)$ with the ground-truth drift $\tilde{b}(Z_t,t)=0.3(80-Z_t)$ on a grid. While the learned drift does not perfectly match the ground truth everywhere, it captures some of the global structure over space-time regions where the CIR process is near steady state (i.e., where $t \gg 0$). These are precisely the space-time regions that are most relevant for sample generation.

\begin{figure}[htbp]
    \centering
    \includegraphics[width=0.6\linewidth]{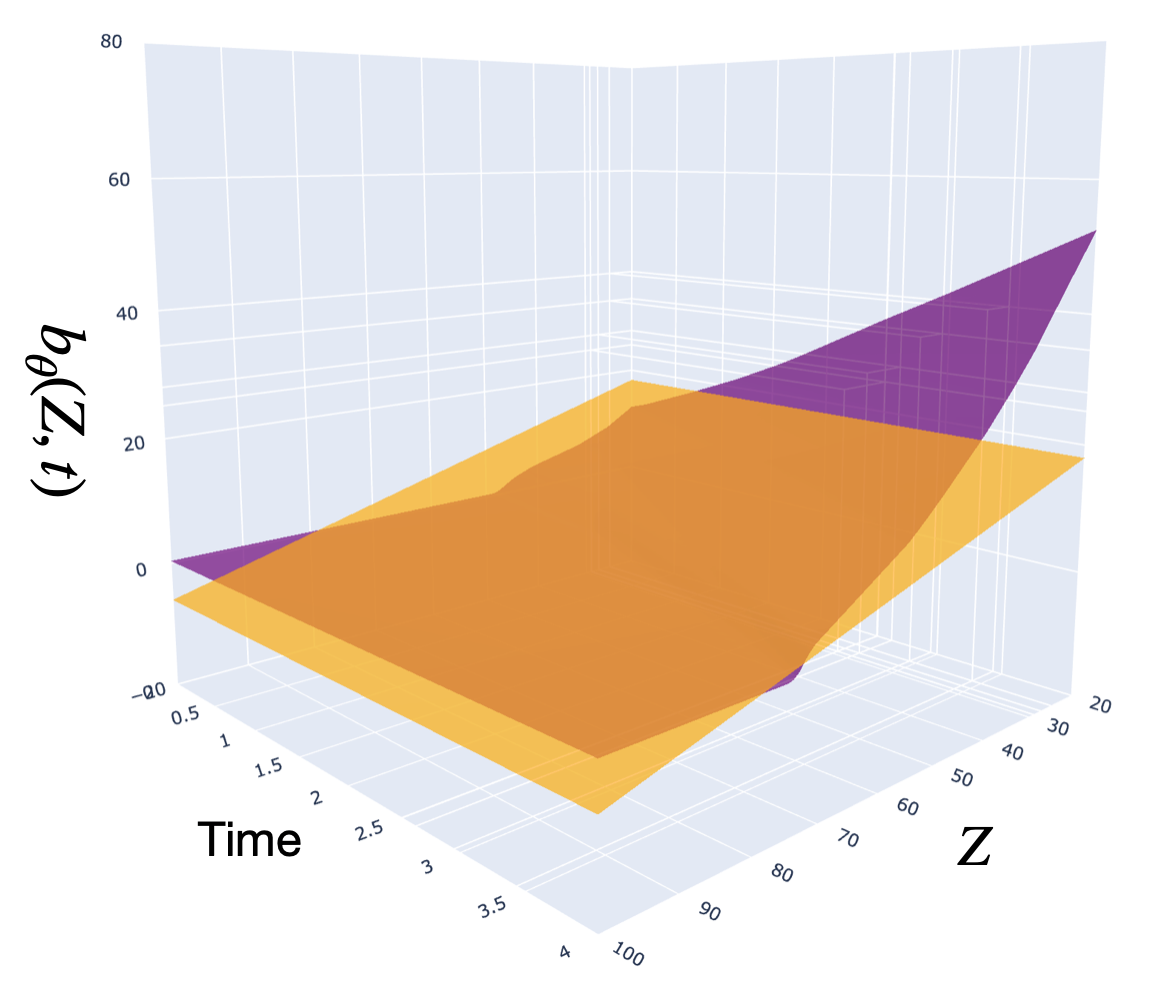}
    \caption{Comparing  learned prior drift $b_\theta(Z_t,t)$ (purple) to the ground truth drift $\tilde{b}(Z_t,t)=0.3(80-Z_t)$ (yellow).}
    \label{fig:prior_drift}
\end{figure}

From the perspective of prediction, 
a more important criterion is the model's generative performance. 
To this end, we generate $128$ Cox process samples using the ground-truth drift $\tilde{b}$ and another $128$ samples using the learned drift $b_\theta$. Figure~\ref{fig:sample_reconstruction} compares the resulting samples in terms of their empirical means and standard deviations. The close agreement between the two sets of statistics indicates that the model captures a representation of the underlying data distribution despite the local deviations shown in Fig.~\ref{fig:prior_drift}.

\begin{figure}[htbp]
    \centering
    \includegraphics[width=0.9\linewidth]{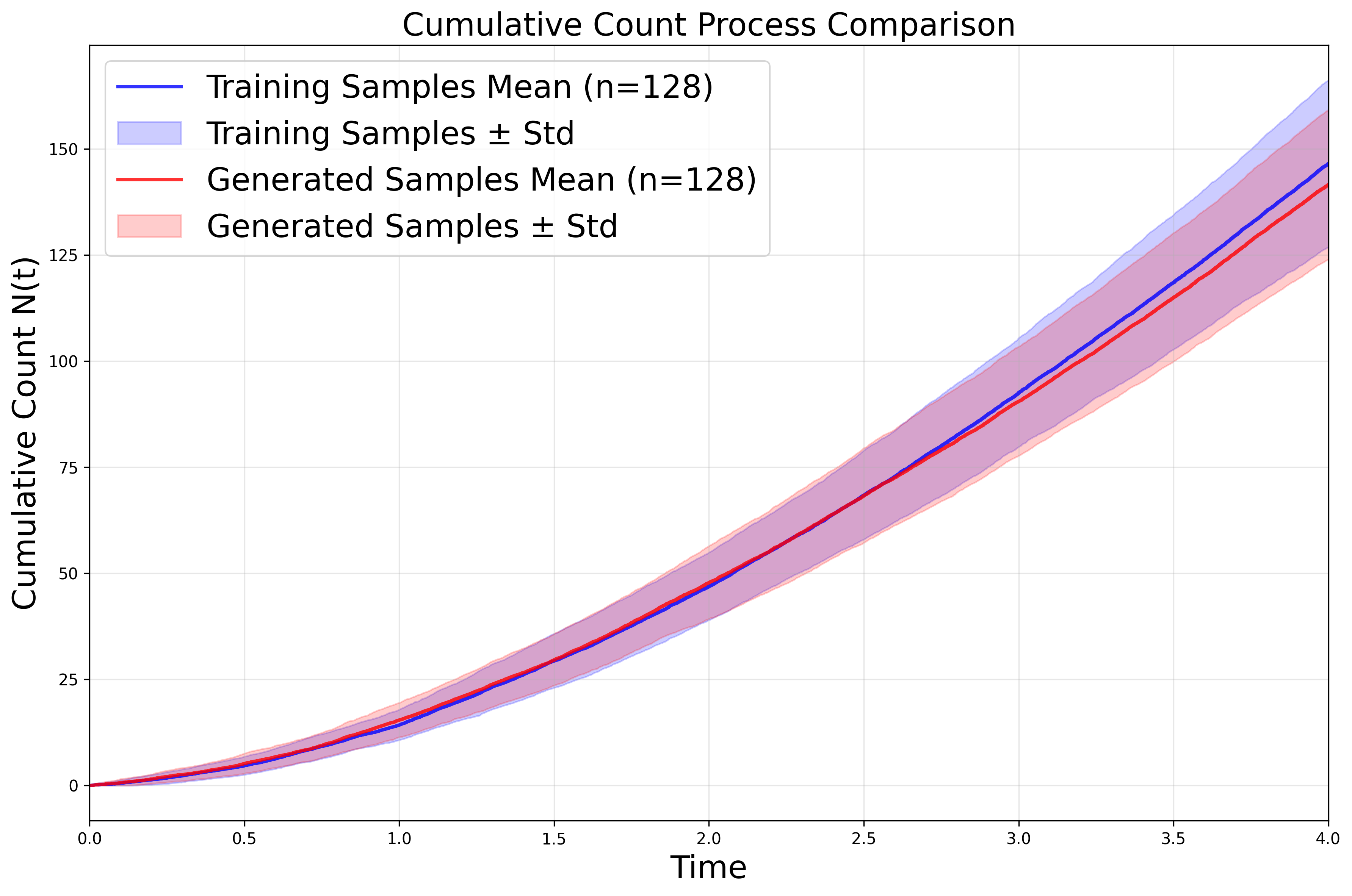}
    \caption{The learned data samples (red) compared to the true data samples (blue).}
    \label{fig:sample_reconstruction}
\end{figure}

\subsection{Amortized Posterior Inference}

To assess the quality of posterior approximation resulting from our amortized variational framework, 
we compare the learned variational posterior against the gold-standard (but computationally intensive) MCMC approximation, run with a burn-in period of 10,000 iterations. 
Fig.~\ref{fig:posterior} shows two representative inference scenarios.

In Fig.~\ref{fig:small_sample_complete}, the posterior is conditioned on the point process sample in the test dataset with the fewest points over $[0,T]$ {in the test dataset}. This is challenging since such extreme samples are very limited in the training dataset. Despite this difficulty, the variational posterior closely tracks the MCMC posterior trajectories.

Fig.~\ref{fig:large_sample_partial} illustrates a more realistic prediction task, where we have access to the point process on $[0,T/2]$, and the goal is to infer posterior dynamics on the unobserved interval $[T/2,T]$. The agreement with the MCMC baseline indicates that the learned drift correction $u_\beta$ effectively incorporates the information provided by the partial observations. 

\begin{figure}[htbp]
    \centering
    
    \begin{subfigure}{\linewidth}
        \centering
        \includegraphics[width=0.9\linewidth]{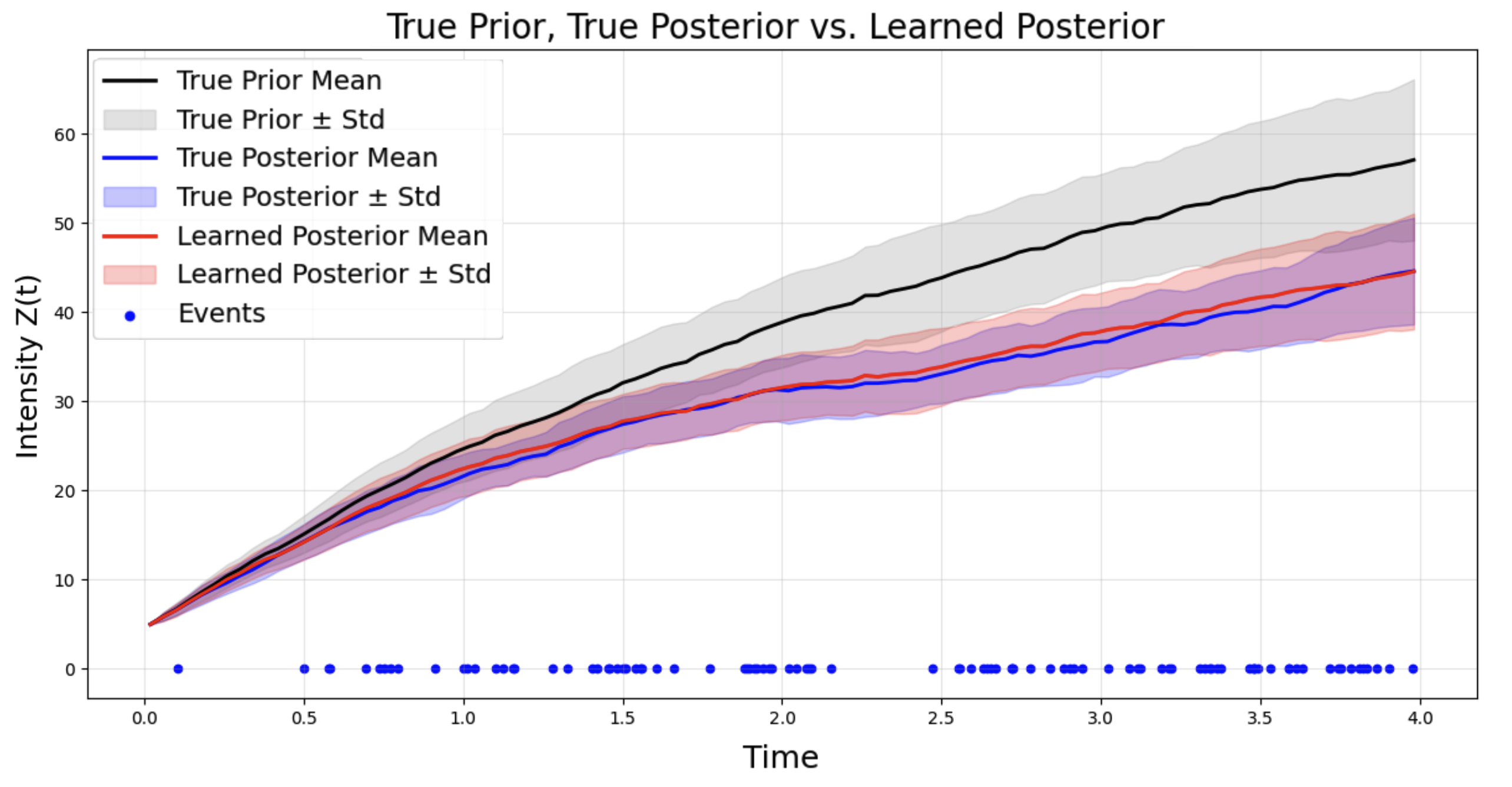}
        \caption{Posterior Inference with complete data $N_{0:T}$.}
        \label{fig:small_sample_complete}
    \end{subfigure}
    \vfill
    \begin{subfigure}{\linewidth}
        \centering
        \includegraphics[width=0.9\linewidth]{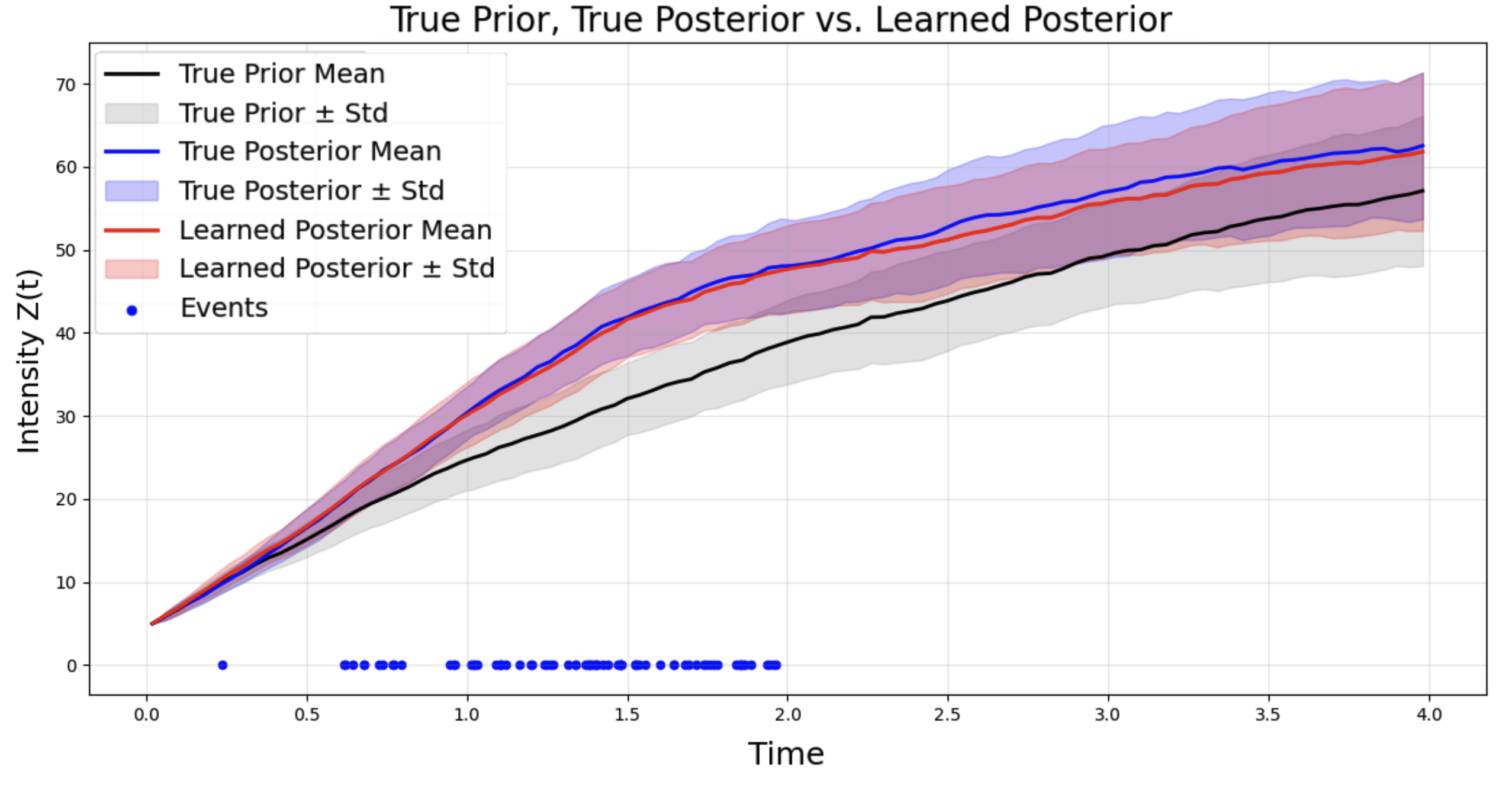}
        \caption{Posterior Inference with partially observed data $N_{0:T/2}$.}
        \label{fig:large_sample_partial}
    \end{subfigure}
    \caption{The learned posterior sample paths (red) are generated using the learned $(b_\theta,u_\beta)$ pair, and the baseline ``true" posterior sample paths are generated using extensive MCMC simulations (blue).}
    \label{fig:posterior}
\end{figure}

\noindent\textbf{\textsc{Does amortization overfit?: }}
A natural question concerns the number of point process realizations (and the observation intervals) necessary for the amortized posterior drift correction $X \mapsto u_\beta(z,t,T', X)$ to reliably generalize.
Even if the prior $b_\theta$ is learned well, an overfitted $u_\beta$ may still yield unreliable posterior predictions for new point process observations. 

To probe this, we measure how closely posterior paths sampled from our amortized posterior $Q_\phi(Z|X)$ match samples from a high-fidelity MCMC approximation of the true posterior $P(Z|X)$, and compare this on training versus testing point process observations.  
As a metric between the resulting path measures, we use the $2$-Wasserstein distance:
\begin{equation}
W(\mu,\nu):=\left(\inf_{\pi\in\Pi(\mu,\nu)}\int d(z,\tilde{z})^2d\pi(Z,\tilde{Z})\right)^{1/2},
\end{equation}
where $\Pi(\mu,\nu)$ denotes the set of couplings of $\mu$ and $\nu$. In our experiments, $Z$ is a latent intensity trajectory on $[0,T]$, and we equip the space with and $L^2$-type metric. Given a time grid $t_{j}=j\Delta t$ for $j=1, ..., M$ (with $M\Delta t=T$), we approximate 
$\hat{d}(Z,\tilde{Z})^2:=\sum_{j=0}^M\Delta t(Z_{t_j}-\tilde{Z}_{t_j})^2.$
For empirical measures $\hat{\mu}=\frac{1}{n}\sum_{i=1}^n\delta_{Z^i}$ and $\hat\nu=\frac{1}{n}\sum_{i=1}^n\delta_{\tilde{Z}^i}$ supported on the $n$ sampled trajectories from $Q_\phi(\cdot|X)$ and $P(\cdot|X)$, respectively, the discrete version of the $2$-Wasserstein distance is computed as
$$\hat{W}(\hat{\mu},\hat{\nu}):=\left(\min_{\gamma\in S_n}\frac{1}{n}\sum_{i=1}^n\hat{d}(Z^{\gamma(i)},\tilde{Z}^i)^2\right)^{1/2},$$
where $S_n$ is the set of all permutations on $\{1,...,n\}$.

\begin{figure}[htbp]
    \centering
    \includegraphics[width=0.6\linewidth]{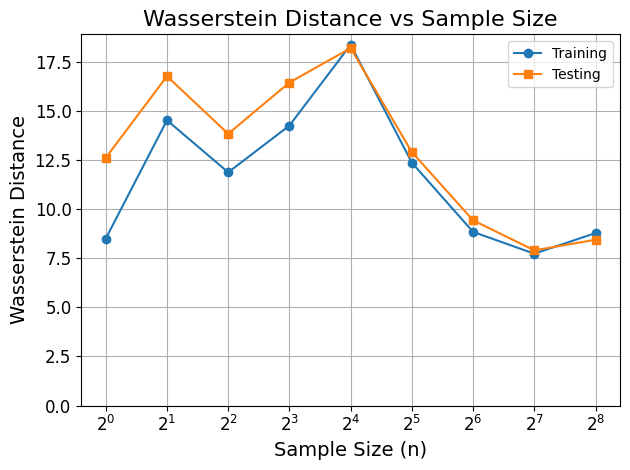}
    \caption{Train vs. test Wasserstein distance between amortized posterior samples and high-fidelity MCMC posterior samples as a function of the training sample size $n$. A train-test gap indicates overfitting of the amortized correction $u_\beta(\cdots, X)$, which vanishes as $n$ goes beyond $16$. }
    \label{fig:amortization_comparison}
\end{figure}

Fig.~\ref{fig:amortization_comparison} shows that for small training sample size (e.g., $n\le 8$), the Wasserstein distance to the high-fidelity MCMC approximation of the posterior is noticeably lower on the training set compared to the testing set, indicating overfitting of the decoder $u_\beta$. As the number of point process observations in the training dataset increases, this discrepancy vanishes, suggesting that amortized posterior inference generalizes reliably to samples outside the training data once the training data size is large enough (e.g., $n\ge 16$).
\subsection{Comparison to EM}

We next compare the proposed variational approach with an approximate expectation-maximization (EM) baseline (implementing Algorithm~\ref{alg:em}) in the same synthetic CIR setting. As introduced previously, the EM method alternates between approximating the posterior of the latent intensity (via MCMC) and maximizing the expected data log-likelihood (via approximate gradient descent). 

\noindent\textbf{\textsc{Learning the Model (Prior): }}
We first evaluate how well each method recovers the ground-truth prior dynamics. As a quantitative metric, we consider the expected $L^2$ deviation between the learned intensity path $Z_t^\theta$ and the ground-truth path $Z_t^{\text{gt}}$:
$
\E{\int_{0}^T\left(Z^\theta_t-Z^{\text{gt}}_t \right)^2\,dt}$,  smaller values indicate better recovery.
This expectation is estimated by simulating both processes using the same set of $64$ i.i.d. Brownian motion realizations. We assume a fixed computational budget for both the EM algorithm and the variational method of five hour run time\footnote{on a M3 MacBook Air with 8GB of RAM}. Table~\ref{tab:prior_comparison} shows the resulting $L^2$ deviations for two different ground-truth drift functions. under the same training time, the variational method achieves comparable (or slightly better) recovery of the prior. Notably, this is achieved while simultaneously learning a posterior drift correction network $u_\beta$, which enables fast amortized inference that's absent from the EM framework. 
\begin{table}[h]
\centering
\begin{tabularx}{\linewidth}{Y Y Y}
\hline
True drift $\tilde{b}$  & EM ($\ell^2$ loss) & VI ($\ell^2$ loss) \\
\hline
$0.3(80-Z_t)$ & 6.159 & 5.792 \\
$0.3(80-Z_t)-5t$ & 6.721 & 4.989 \\
\hline
\end{tabularx}
\caption{Comparison of the learned prior quality in terms of $\ell^2$ deviation of the intensity paths. }
\label{tab:prior_comparison}
\end{table}

\noindent\textbf{\textsc{Inference Time:}}
We compare  posterior inference `efficiency' between the two methods. Specifically, rather than a fixed computational budget, we now fix a desired accuracy of the posterior estimate and measure the time required to achieve that. The quality of the posterior samples is quantified with the posterior predictive log-likelihood:
\begin{equation}
\E[X\sim\mugt]{{\E[Z\sim Q(\cdot|X)]{\sum_{T'<\tau_i\le T}\log Z_{\tau_i}-\int_{T'}^TZ_t\,dt}}},
\end{equation}
which evaluates how well posterior samples explain future observations. 

For the variational method, samples from $Q(Z|X) \equiv Q_\phi(Z|X)$ are obtained by simulating the posterior SDE, Eq.~\eqref{eq:posterior_sde}, using the learned $(b_\theta,u_\beta)$ pair. In contrast, we use a Metropolis-Hastings sampler for the EM baseline. For each prediction horizon $[T',T]$, we tune the number of MCMC steps such that both methods get comparable posterior predictive log-likelihoods. We then record the total time required to generate $32$ posterior sample paths conditioned on all $128$ testing point process observations.

\begin{table}[h]
\centering
$\tilde{b}(Z_t,t)=0.3(80-Z_t)$\\
\begin{tabularx}{\linewidth}{Y Y Y Y}
\hline
Horizon & Likelihood & MCMC & VI \\
\hline
$[0,T]$ & 398.0 & 17m 38.5s & 39.9s \\
$[T/4,T]$ & 370.1 & 12m 43.8s & 9.6s \\
$[T/2,T]$ & 288.4 & 14m 3.3s & 19.6s \\
$[3T/4,T]$ & 162.6 & 15m 49.8s & 29.7s \\
\hline
\end{tabularx}
\caption{Comparison of computation time used for posterior inference on the testing dataset. 
The data is generated with the ground truth drift $\tilde{b}(Z_t,t)=0.3(80-Z_t)$.}
\label{tab:posterior_time_hom}
\end{table}

\begin{table}[h]
\centering
$\tilde{b}(Z_t,t)=0.3(80-Z_t)-5t$\\
\begin{tabularx}{\linewidth}{Y Y Y Y}
\hline
Horizon & Likelihood & MCMC & VI \\
\hline
$[0,T]$ & 249.9 & 16m 17.7s & 38.2s \\
\hline
\end{tabularx}
\caption{Comparison of computation time with time-dependent $\tilde{b}(Z_t,t)=0.3(80-Z_t)-5t$.}
\label{tab:posterior_time_inhom}
\end{table}

Tables~\ref{tab:posterior_time_hom} and~\ref{tab:posterior_time_inhom} summarize the results for time-homogeneous and time-inhomogeneous ground-truth drifts, respectively. Across all horizons, the variational method is one to two orders of magnitude faster than MCMC-based posterior sampling while achieving similar predictive log-likelihoods. This efficiency gap widens as $T'$ decreases, demonstrating the computational advantages of amortization in this setting.

\subsection{US Bank Dataset}\label{sec:us_bank}

Finally, we evaluate the proposed method on a real-world dataset of call records from a large U.S.-based bank call center\footnote{https://seelab.net.technion.ac.il/data/, or see Appendix \ref{sec:call_data}}. The call record counts demonstrate strong dyadic or time-of-day variation and over-dispersion (~\citep{Brown2005}). The over-dispersion of the counts immediately imply that the process is  well-modeled by a Cox process, and indeed~\citep{Brown2005} describe time-binned models for estimating a random daily-effect stochastic model of the intensity. The dataset contains the number of arrivals at each minute of the day, which are turned into exact arrival times by evenly spreading those arrivals across the minute. We use data from $128$ Monday's as the training set. The arrivals are then thinned independently with thinning probability $0.001$ (keep each arrival with probability $0.001$ independently) for training stability. 

The model is trained using the same optimization procedures as in the synthetic setting, with the same hyperparameters. The training time is constrained to $5$ hours as with the synthetic experiments. Since the true prior is unknown, we directly compare samples of the reconstructed samples to the entire training data $(n=128)$ in Fig.~\ref{fig:sample_reconstruction_usbank}. Qualitatively, the learned latent intensity mean trajectory captures the main temporal patterns observed in the dataset, including peak periods, but the standard deviation trajectories do not closely match the data. This is most likely due to the fact that the diffusion coefficient is fixed to $\sigma(Z_t,t)=\sqrt{Z_t}$, and one natural improvement of the proposed methodology is to model the diffusion with another neural network $\sigma_\alpha$. The entire methodology remains unchanged except for changes in the gradient computations.

\begin{figure}[htbp]
    \centering
    \includegraphics[width=0.9\linewidth]{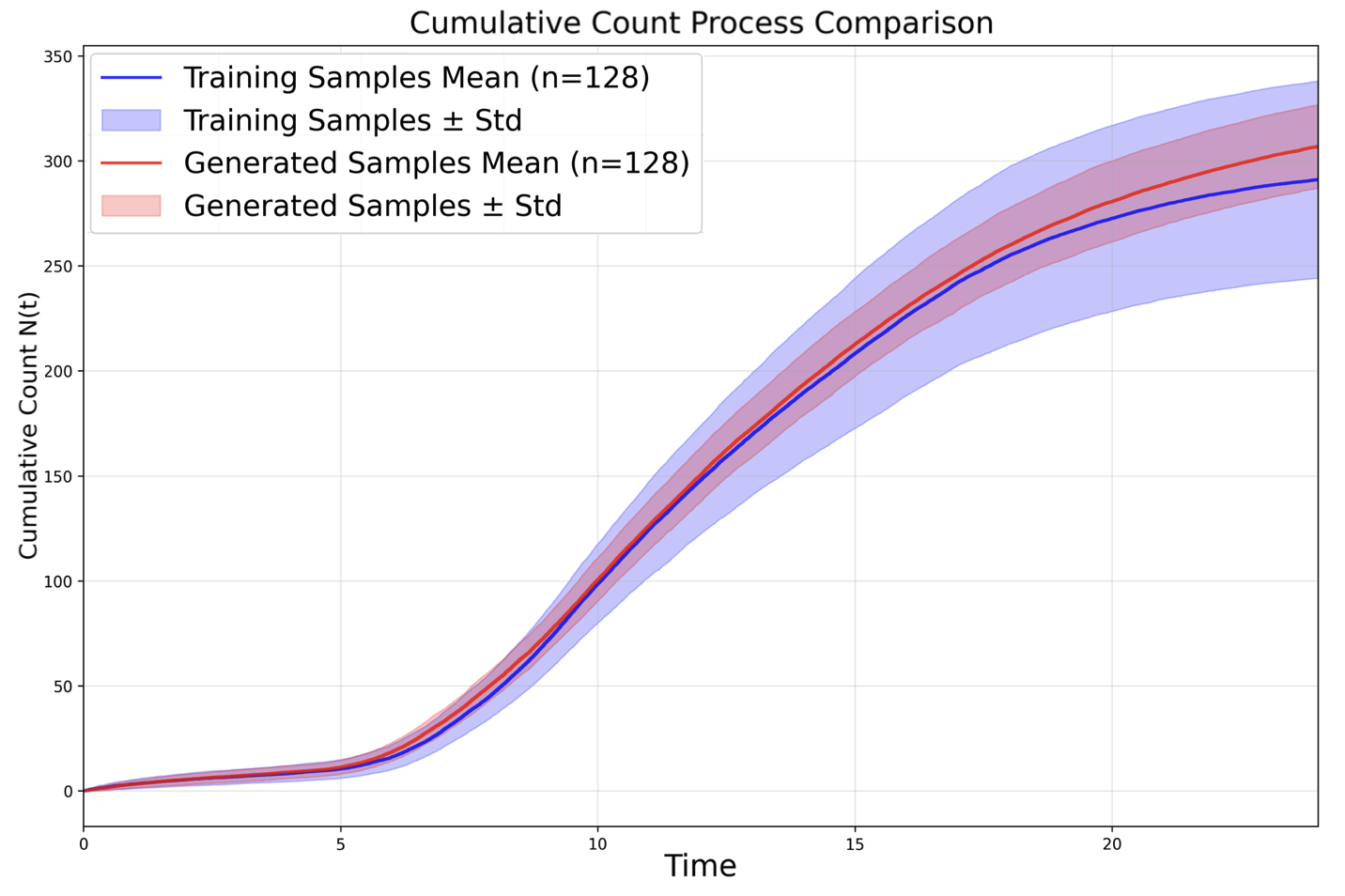}
    \caption{The learned data samples (red) compared to the true data samples (blue).}
    \label{fig:sample_reconstruction_usbank}
\end{figure}

\section{Conclusion}

We introduce Neural Diffusion Intensity Models, a variational framework for Cox process models where the latent intensity evolves as a neural SDE. Our key theoretical contribution leverages enlargement of filtrations to show that conditioning on point process observations preserves the diffusion structure of the latent intensity with an explicit drift correction — establishing a conjugacy between our neural SDE priors and the Poisson likelihood. This ensures the variational family contains the true posterior, so ELBO maximization coincides with likelihood maximization under sufficient capacity. Unlike existing frameworks~\cite{Opper2019,tzen2019neural}, where the posterior drift is characterized only as the solution to a stochastic control problem, ours is available in closed form.

We propose an amortized inference framework with a DeepSets-inspired architecture for the posterior correction drift, avoiding the repeated MCMC simulations of classical EM methods. Experiments on synthetic and real data show accurate recovery of prior dynamics with orders-of-magnitude speedups over MCMC-based posterior inference. Our enlargement-of-filtrations methodology is broadly applicable and potentially extensible beyond Cox models.

\bibliographystyle{plain}
\bibliography{references}

\newpage
\appendix
\section*{Supplementary Material}
\addcontentsline{toc}{section}{Supplementary Material}
\section{Measure-Theoretic Foundations of the Neural SDE Cox Model}~\label{sec:neural-cox-deets}
\label{app:measure-theory}

We provide precise conditions under which the path measures in the Neural SDE Cox model (Section~\ref{sec:neural-cox}) are well-defined. Throughout, let $\mathcal{Z} := C([0,T], \mathbb{R}_{\geq 0})$ denote the space of non-negative continuous paths on $[0,T]$ equipped with the sup-norm topology and its Borel $\sigma$-algebra $\mathcal{B}(\mathcal{Z})$, and let $\mathcal{X} = \cup_{k \in \mathbb N} \mathbb R^k$ denote the space of locally finite counting measures on $[0,T]$ equipped with the vague topology and its Borel $\sigma$-algebra $\mathcal{B}(\mathcal{X})$.

\paragraph{Existence and uniqueness of the SDE solution.}
We require the drift and diffusion coefficients of the SDE~\eqref{eq:sde} to satisfy the following standard conditions:
\begin{enumerate}[label=(\textbf{A\arabic*})]
    \item \label{cond:lipschitz} \textit{Global Lipschitz continuity.} There exists a constant $K > 0$ such that for all $x, y \in \mathbb{R}$ and $t \in [0,T]$,
    \[
    |b_\theta(x,t) - b_\theta(y,t)| + |\sigma(x,t) - \sigma(y,t)| \leq K|x - y|.
    \]
    \item \label{cond:growth} \textit{Linear growth.} There exists a constant $C > 0$ such that for all $x \in \mathbb{R}$ and $t \in [0,T]$,
    \[
    |b_\theta(x,t)| + |\sigma(x,t)| \leq C(1 + |x|).
    \]
    \item \label{cond:measurability} \textit{Joint measurability.} The functions $b_\theta(\cdot, \cdot)$ and $\sigma(\cdot, \cdot)$ are jointly Borel measurable in $(x, t)$.
\end{enumerate}
Under \ref{cond:lipschitz}--\ref{cond:measurability}, the SDE~\eqref{eq:sde} admits a unique strong solution for each $\theta \in \Theta$ \citep[Theorem~5.2.1]{oksendal2003stochastic}. This solution induces a well-defined probability measure $P_\theta(Z)$ on $(\mathcal{Z}, \mathcal{B}(\mathcal{Z}))$ via the pushforward of the Wiener measure through the It\^o map.

\paragraph{Non-negativity of the intensity process.}
Since $Z$ serves as an intensity, we additionally require:
\begin{enumerate}[label=(\textbf{A\arabic*}), resume]
    \item \label{cond:nonneg} \textit{Non-negativity.} $Z_t \geq 0$ almost surely for all $t \in [0,T]$.
\end{enumerate}
This condition is not implied by \ref{cond:lipschitz}--\ref{cond:measurability} and must be enforced through the model structure. For instance, if $\sigma(z,t) = \sigma_0 \sqrt{z}$ (as in the CIR model), the Feller condition $2b_\theta(0,t) \geq \sigma_0^2$ for all $t$ ensures that the boundary at zero is unattainable. Alternatively, one may model $\log Z_t$ as the SDE solution and recover $Z_t$ by exponentiation, guaranteeing positivity.

\paragraph{The Poisson kernel as a Markov kernel.}
Given a non-negative intensity path $Z \in \mathcal{Z}$, the conditional law $P(X \mid Z)$ is that of an inhomogeneous Poisson process with rate function $(Z_t)_{t \in [0,T]}$. Its Radon--Nikodym derivative with respect to the unit-rate Poisson measure $P_0$ is
\begin{equation}
\label{eq:poisson-rn}
\frac{dP(X \mid Z)}{dP_0} = \exp\!\left(\int_0^T \log Z_t \, dN_t - \int_0^T (Z_t - 1) \, dt \right),
\end{equation}
where $N_t$ denotes the counting process associated with $X$. We require:
\begin{enumerate}[label=(\textbf{A\arabic*}), resume]
    \item \label{cond:integrability} \textit{Integrability.} $\int_0^T Z_t \, dt < \infty$ almost surely.
    \item \label{cond:positivity} \textit{Strict positivity at event times.} $P_\theta(Z_t > 0 \text{ for all } t \in [0,T]) = 1$.
\end{enumerate}
Condition~\ref{cond:integrability} is automatically satisfied when $Z$ has continuous paths on the compact interval $[0,T]$, but we state it explicitly for completeness. Condition~\ref{cond:positivity} ensures that $\log Z_t$ in~\eqref{eq:poisson-rn} is well-defined at event times; it is implied by the Feller condition in the CIR setting or by the log-normal construction. Under \ref{cond:integrability}--\ref{cond:positivity}, the map $Z \mapsto P(X \mid Z)$ defines a Markov kernel from $(\mathcal{Z}, \mathcal{B}(\mathcal{Z}))$ to $(\mathcal{X}, \mathcal{B}(\mathcal{X}))$, since the Poisson likelihood~\eqref{eq:poisson-rn} is a measurable functional of the intensity path.

\paragraph{Well-definedness of the joint and marginal measures.}
Under conditions \ref{cond:lipschitz}--\ref{cond:positivity}, the joint measure
\[
P_\theta(X, Z) = P(X \mid Z) \, P_\theta(Z)
\]
is well-defined on the product $\sigma$-algebra $\mathcal{B}(\mathcal{X}) \otimes \mathcal{B}(\mathcal{Z})$ by the standard Markov kernel construction \citep[Chapter~3]{kallenberg2021foundations}. The marginal observation law
\[
P_\theta(X) = \int_{\mathcal{Z}} P(X \mid Z) \, dP_\theta(Z)
\]
is then a well-defined probability measure on $(\mathcal{X}, \mathcal{B}(\mathcal{X}))$.

\paragraph{Regularity in $\theta$.}

For gradient-based optimization via the pathwise (``differentiate then simulate'') method~\citep{glasserman1991gradient,gobet2005sensitivity}, we further require:
\begin{enumerate}[label=(\textbf{A\arabic*}), resume]
    \item \label{cond:differentiability} \textit{Differentiability in parameters.} The map $\theta \mapsto b_\theta(x,t)$ is continuously differentiable for all $(x,t)$, and $\nabla_\theta b_\theta$ satisfies conditions \ref{cond:lipschitz}--\ref{cond:growth} uniformly in $\theta$ over compact subsets of $\Theta$.
\end{enumerate}
Under \ref{cond:differentiability}, the augmented system of SDEs for $(Z_t, \partial_\theta Z_t, \partial_\beta Z_t)$ admits a unique strong solution, ensuring that the pathwise sensitivity processes are well-defined and that the resulting Monte Carlo gradient estimators are unbiased (up to time-discretization error).

\section{Enlargement of Filtrations}\label{sec:eof}
In this section, we will work with a \textit{filtered probability space} $(\Omega, \F,\mathbb{F},\P)$ where the probability measure $\P$ encodes our ``beliefs'', and the filtration---collection of $\sigma$-algebras---$\mathbb{F}=(\F_t)_{t\ge0}$ represents the information that's available to us at any time. A stochastic process $(X_t)_{t\ge0}$ is said to be \textit{adapted} to $\mathbb{F}$ if $X_t$ is $\mathcal{F}_t$-measurable for all $t$. That is, the value of $X_t$ can be determined exactly by the information available at time $t$. The mathematical framework that studies how stochastic processes behave under such changes in information is called {Enlargement of Filtrations}, which has been studied extensively in the probability literature; see \citep{Grigorian2023, Jacod2006, Jeanblanc2009}.

\subsection{Martingales}
Let $(\Omega, \F,\mathbb{F}=(\F_t)_{t\ge0},\P)$ be a filtered probability space, and let $M_t$ be an $\mathbb{F}$-adapted stochastic process with $\mathbb{E}\abs{M_t}<\infty$ for all $t$. If 
$$\E{M_t|\F_s}=M_s$$
for all $0\le s< t$, then $M_t$ is said to be an $\mathbb{F}$-martingale. Intuitively, a martingale represents a ``fair game": given the current information, its expected future value is just the present value. A process $X_t$ is a \textit{semimartingale} if it can be decomposed as
$$X_t=M_t+A_t$$
where $M_t$ is a local martingale (which behaves like a martingale up to random stopping times)\footnote{A process $M_t$ is a local martingale if there exists an increasing sequence of stopping times $\tau_n\nearrow\infty$ such that each stopped process $M_{t\wedge\tau_n}$ is a martingale.} and $A_t$ is an $\mathbb{F}$-adapted finite-variation process. Semimartingales form the largest class of stochastic processes for which we can talk about It\^o calculus. 

\subsection{Enlarging the filtration}
$X_t$ may be a martingale with respect to some filtration $\mathbb{F}$, but not necessarily when $\mathbb{F}$ is ``enlarged" with extra information $\zeta$. Describing $X_t$ in the presence of this information is the goal of Enlargement of Filtrations. 

\subsubsection{Setup}
There are two main types of ``enlargement":
\begin{itemize}
\item \textbf{Initial enlargement:} all secret information is available at time $0$.
\item \textbf{Progressive enlargement:} information is gradually revealed over time.
\end{itemize}

We focus on the first case, which will be related to our variational problem. Let $\mathbb{F} = (\mathcal{F}_t)_{t \ge 0}$ be the original filtration, and let $\zeta$ be an $\mathcal{F}$-measurable random variable taking values in some measurable space $E$.  
We define the \emph{enlarged filtration} $\mathbb{G} = (\mathcal{G}_t)_{t \ge 0}$ by
$$
\mathcal{G}_t = \mathcal{F}_t \vee \sigma(\zeta),
$$
where $\vee$ denotes the smallest $\sigma$-algebra containing elements from $\mathcal{F}_t$ and $\sigma(\zeta)$. Intuitively, $\mathbb{G}$ describes a world in which the value of $\zeta$ is known from the start---one has “insider” information about $\zeta$ at time $0$.

\subsubsection{The $\H$ and $\H'$ Hypotheses}
The question is how martingales behave under a filtration enlargement. 

If every $\mathbb{F}$-martingale remains a $\mathbb{G}$-martingale, we say that $\mathbb{F}$ is \textit{immersed} in $\mathbb{G}$, or that the pair $(\mathbb{F}, \mathbb{G})$ satisfies the \textbf{$\H$-hypothesis}. This is quite a strong condition---informally, it means that the extra information carried by $\zeta$ does not interfere with the “fairness” of any $\mathbb{F}$-martingale.

A weaker but more flexible condition is the \textbf{$\H'$-hypothesis}: every $\mathbb{F}$-martingale remains a $\mathbb{G}$-\textit{semimartingale}. Under this assumption, It\^o calculus remains valid, but an additional drift term may appear when expressing the $\mathbb{F}$-martingale in the $\mathbb{G}$ filtration. This is what's going under the hood when we enlarge the filtration with the observed point process $N_{0:T'}$.

\subsubsection{Jacod's condition}
Among several sufficient conditions ensuring the $\mathcal{H}'$-hypothesis, \emph{Jacod's condition} is often the most convenient in practice. The following statements (theorem and lemma) are adapted from \citep{Grigorian2023}.
\begin{theorem}[Jacod's condition]\label{thm:jacod_condition}
Let $\zeta$ be a random element in a standard Borel space $(E,\mathcal{E})$, and let $Q_t(\omega, dx)$ be the regular conditional distribution of $\zeta$ given $\F_t$ for all $t\ge0$. Suppose that there exists a positive $\sigma$-finite\footnote{A measure is called $\sigma$-finite if it can be written as a countable union of finite measures.} measure $\eta$ on $(E,\mathcal{E})$ such that 
$$Q_t(\omega, dx)\ll\eta(dx)\text{ a.s., }\forall t\ge0.$$
Then $\H'$ holds.
\end{theorem}
Just knowing that a process remains a semimartingale is usually not enough, we typically would like the semimartingale decomposition in the enlarged filtration\footnote{We use $\mathbb{F}^{\sigma(\zeta)}$ to denote the right-continuous version of the enlarged filtration $\mathbb{F}\lor \sigma(\zeta)$.} $\mathbb{F}^{\sigma(\zeta)}$. To do so, we need the following lemma and theorem.

\begin{lemma}
Under $\mathcal{J}$, there exists a nonnegative $\mathcal{O}(\mathbb{F})\times\mathcal{B}(\bar{\mathbb{R}})$-measurable function 
$$\Omega\times \mathbb{R}_+\times\bar{\mathbb{R}}\ni (\omega, t,x)\mapsto p_t(\omega, x)$$
cadlag in $t$ such that 
\begin{itemize}
\item[(i)]
$Q_t(\omega,dx)=p_t(\omega,x)\eta(dx)$ for every $t\ge0$,
\item[(ii)]
for each $x\in\bar{\mathbb{R}}$, the process $(p_t(x))_{t\ge0}$ is an $\mathbb{F}$-martingale,
\item[(iii)]
$p(x)>0$ and $p_-(x)>0$ on\footnote{The double square bracket is used for intervals with random boundary points. } $\llbracket0,\zeta^x\llbracket$ and $p(x)=0$ on $\llbracket\zeta^x,\infty\llbracket$, where $\zeta^x:=\inf\{t:p_{-}(x)=0\}$.
\end{itemize} 
\end{lemma}
\begin{theorem}\label{thm:decomposition}
Suppose that the random element $\zeta$ satisfies $\mathcal{J}$. If $X$ is an $\mathbb{F}$-local martingale, the process $\tilde{X}$ defined as
$$\tilde{X}_t:=X_t-\int_0^t\frac{1}{p_{s-}(\zeta)}d\langle{X,p(u)\rangle}^\mathbb{F}_s|_{u=\zeta}, \quad t\le T\quad (*)$$
is an $\mathbb{F}^{\sigma(\zeta)}$-local martingale. 
\end{theorem}

\subsubsection{Example: Brownian bridge via initial enlargement}

Consider a standard Brownian motion $B=(B_t)_{t\ge 0}$ on $(\Omega,\mathcal{F},\mathbb{F},\mathbb{P})$ with its natural filtration ($\F_t=\sigma(X_s,0\le s\le t)$). Suppose we enlarge $\mathbb{F}$ with the information given by $\zeta:=B_T$, i.e.,
$$
\mathcal{G}_t = \mathcal{F}_t \vee \sigma(B_T), \qquad \mathbb{G}=(\mathcal{G}_t)_{t\ge 0}.
$$
As always, $\mathbb{G}$ is assumed to be right-continuous by taking 
$$\mathcal{G}^+_t=\bigcap_{s>t}\mathcal{G}_s.$$

\paragraph{Step 1: Verifying Jacod's condition.}
For $t<T$, the conditional law of $B_T$ given $\mathcal{F}_t$ is Gaussian:
\[
B_T \mid \mathcal{F}_t \sim \mathcal{N}\big(B_t,\; T-t\big).
\]
Hence there exists a dominating measure (the Lebesgue measure) $\eta(dx)=dx$. $Q_t(\omega,dx)$ has a density with respect to Lebesgue measure: $Q_t(\omega,dx) \;=\; p_t(\omega,x)\,dx$, with
$$
p_t(x) = \frac{1}{\sqrt{2\pi (T-t)}}\exp\!\Big(-\frac{(x-B_t)^2}{2(T-t)}\Big),\quad 0\le t<T.
$$
Thus $Q_t(\cdot,dx)\ll \eta (dx)$ a.s. for each $t<T$, so Jacod's condition holds on $[0, t)$.  (On $[T,\infty)$, the conditional law is degenerate at $B_T$.)

\paragraph{Step 2: Computing the semimartingale decomposition.}
We apply Theorem \ref{thm:decomposition} with $X=B$. To do so, we first compute the $\mathbb{F}$-predictable covariation $\langle B, p(x)\rangle^\mathbb{F}$.  
By It\^o’s formula, we can get 
$$
dp_t(x) = \partial_{B_t} p_t(x)\, dB_t
\;=\; \frac{x-B_t}{T-t}\, p_t(x)\, dB_t, \qquad t<T.
$$
Hence the predictable covariation with $B$ is just
$$
d\big\langle B, p(x)\big\rangle_t^{\mathbb{F}} = \frac{x-B_t}{T-t}\, p_t(x)\, dt, 
\qquad 0\le t<T.
$$
Plugging this into $(*)$ with $x=\zeta=B_T$, we obtain from Theorem \ref{thm:decomposition} that the process
$$
\tilde{B}_t:=B_t-\int_0^t\frac{B_T-B_s}{T-s}\,ds, \qquad0\le t<T
$$
is a $\mathbb{G}$-local martingale. Since $B$ is continuous with quadratic variation $[B]_t=t$, it follows that $[\tilde{B}]_t=t$ as well, and by L\'evy’s characterization $\tilde{B}$ is a $\mathbb{G}$-Brownian motion on $[0,T)$.  

Equivalently, $B$ admits the $\mathbb{G}$-semimartingale (indeed, SDE) representation
$$
dB_t = d\tilde{B}_t \;+\; \frac{B_T - B_t}{T-t}\, dt, 
\qquad 0\le t<T,
$$
where $\tilde{B}$ is a $\mathbb{G}$-Brownian motion. This is precisely the \emph{Brownian bridge} drift toward the terminal pinning value $B_T$.

\subsubsection{Proof of Theorem 2.1: Intensity process under initial enlargement}~\label{sec:intense-enlarge}
Now we come back to the setting in the main body of the paper: a nonnegative stochastic process $Z$ has an SDE representation in the natural filtration $\mathbb{F}$:
$$dZ_t=b(Z_t,t)\,dt+\sigma(Z_t,t)\,dB_t, \quad Z_0=z_0.$$
For a Cox process observation $N_{0:T}$, if we consider the enlarged filtration $\mathcal{G}_t=\bigcap_{s>t}(\F_s \lor \sigma(N_{0:T}))$, does $Z$ remain a semimartingale in $\mathbb{G}=(\mathcal{G}_t)_{0\le t\le T}$? In this case, we choose the reference measure to be the unit rate Poisson process so that the Radon-Nikodym process derivative between the conditional measure and the reference measure is 
\begin{equation}
p_t(\omega,x)=\frac{Q_t(\omega,dx)}{\eta(dx)}=\exp\left(\int_0^t\log Z_s\,dN_s-\int_0^t(Z_s-1)\,ds\right)\cdot \beta_t(dx),
\end{equation}
where 
$$\beta_t(dx)=\E{\exp\left(\int_t^T\log Z_s\,dN_s-\int_t^T(Z_s-1)\,ds\right)\Bigg|\F_t}.$$
Assuming $Z_s$ is almost surely positive for all $t\in[0,T]$, then the Radon-Nikodym derivative exists and Jacod's condition is satisfied. To compute the drift correction term, let $h(Z_t,t)=\beta_t(dx)$, where 
$$h(z,t)=\E{\exp\left(\int_t^T\log Z_s\,dN_s-\int_t^T(Z_s-1)\,ds\right)\Bigg|Z_t=z}.$$
By Theorem~\ref{thm:decomposition}, 
$$\tilde{B}_t=B_t-\int_0^t\frac{1}{p_s}d\langle{B,p(u)\rangle}^{\mathbb{F}}_s|_{u=N_{0:T}},t\le T$$
is an $\mathbb{F}^{\sigma(N_{0:T})}$-Brownian motion. It remains to compute the predictable covariation $d\langle{B,p(u)\rangle}_s$:
\begin{align*}
\frac{1}{p_{s-}}d\langle{B,p(u)\rangle}_s&=\frac{1}{h_{s-}}d\langle{B,h\rangle}_s\\
&=\frac{1}{h_s}\sigma\sqrt{Z_t}\frac{\partial}{\partial Z_t}h(Z_t,t)\,dt\\
&=\sigma\sqrt{Z_t}\frac{\partial}{\partial Z_t}\log h(Z_t,t)\,dt
\end{align*}
Altogether, we have 
\begin{align*}
d\tilde{B}_t&=dB_t-\sigma\sqrt{Z_t}\frac{\partial}{\partial Z_t}\log\E{\exp\left(-\int_t^TZ_s\,ds\right)\prod_{\tau_i\in (t,T]}Z_{\tau_i}\Bigg|Z_t}dt.
\end{align*}

Since we want to do inference on the interval $(T',T]$ with the information $N_{0:T'}$ sometimes, we take another look to what happens when we enlarge the filtration $\mathbb{G}$ with $N_{0:T'}$. Take $\eta(dx)$ to be the measure on $D([0,T'], \mathbb{N})$ describing the unit rate Poisson process on $[0, T']$. Then the Radon-Nikodym derivative between the conditional measure and this base measure is
$$p_t(\omega,x)=\frac{Q_t(\omega,dx)}{\eta(dx)}=\exp\left(\int_0^{t\wedge T'}\log Z_s\,dN_s-\int_0^{t\wedge T'}(Z_s-1)\,ds\right)\beta_t(dx),$$
where
$$\beta_t(dx)=\E{\exp\left(\int_{t\wedge T'}^{T'}\log Z_sdN_s-\int_{t\wedge T'}^{T'}(Z_s-1\,ds) \right)\Bigg| \F_t}.$$
Doing similar calculations as before, we get
$$d\tilde{B}_t=dB_t-\mathbf{1}_{\{t\le T'\}}\sigma(Z_t,t)\frac{\partial}{\partial Z_t}\log\E{\exp\left(-\int_t^{T'}Z_s\,ds\right)\prod_{\tau_i\in (t,T']}Z_{\tau_i}\Bigg|Z_t}dt.$$
This means that in the enlarged filtration, the intensity process can be described as
\begin{align*}
dZ_t&=b(Z_t,t)\,dt+\sigma(Z_t,t)\left(d\tilde{B}_t \right. \\ & \qquad\qquad\qquad \left. +\mathbf{1}_{\{t\le T'\}}\sigma(Z_t,t)\frac{\partial}{\partial Z_t}\log\E{\exp\left(-\int_t^{T'}Z_s\,ds\right)\prod_{\tau_i\in (t,T']}Z_{\tau_i}\Bigg|Z_t}\right)dt\\
&=\left[b(Z_t,t)+\mathbf{1}_{\{t\le T'\}}\sigma(Z_t,t)^2\frac{\partial}{\partial Z_t}\log\E{\exp\left(-\int_t^{T'}Z_s\,ds\right)\prod_{\tau_i\in (t,T']}Z_{\tau_i}\Bigg|Z_t}\right]\,dt\\&\qquad\qquad\qquad\qquad\qquad\qquad\qquad\qquad\qquad\qquad\qquad\qquad+\sigma(Z_t,t)\,d\tilde{B}_t,
\end{align*}
which is the form appearing in Theorem~\ref{thm:jacod}. Note that $\tilde{B}$ is a standard Brownian motion with respect to the enlarged filtration, so numerically simulating the above SDE with Euler scheme simply involves replacing $d\tilde{B}_t$ with standard Gaussian increments.

\section{Expectation Maximization and Gradient Computations}\label{sec:em+vi}
In this section, we introduce the general expectation-maximization scheme. Then, we look at an approximate version as applied to our problem setting, with a focus on gradient computations in the M-step. 

\subsection{The Expectation-Maximization Algorithm}\label{sec:em}
First, we restate the variational lower bound in Eq.~\eqref{eq:elbo} for the log-likelihood:
\begin{align}\label{eq:elbo_2}
\log P_\theta(X)\ge \E[Q(Z)]{\log P(X|Z)}-\KL{Q(Z)||P_\theta(Z)}.
\end{align}
The expectation-maximization scheme maximizes this lower bound by iterating between two steps:
\begin{itemize}
    \item \textbf{Expectation}-step\\
    Find $Q(Z)$ that tightens the inequality in Eq.~\eqref{eq:elbo_2}
    \item \textbf{Maximization}-step\\
    Maximize the right-hand side of Eq.~\eqref{eq:elbo_2} with respect to $\theta$. 
\end{itemize}
One can show that the $Q(Z)$ that tightens the inequality is exactly the true posterior, i.e., 
$$Q(Z)=P_\theta(Z|X).$$ This procedure is shown in details in Algorithm~\ref{alg:em}. 

\begin{algorithm}[h]
\caption{Expectation-Maximization}
\label{alg:em}
\KwIn{$n$ observed samples $\{X^i\}_{i=1}^n$}
\textbf{Initialize} with arbitrary $\theta^{(0)}$\\
\For{$t\gets 0$ \KwTo $\infty$}{
  \textbf{E}xpectation: Compute for all $X^i$\\ $Q^i(Z) = P_{\theta^{(t)}}(Z|X^i)$ \;

 \textbf{M}aximization: \\
$\theta^{(t+1)} \in \argmax_\theta \frac{1}{n}\sum_{i=1}^n\left(\E[Q^i(Z)]{\log P(X^i|Z)}-\KL{Q^i(Z)||P_\theta(Z)}\right)$\;
}
\Return $Q^\infty$\;
\end{algorithm}

\subsection{Same algorithm, adapted to our problem}
In our problem setting (and many others), it is hard to compute the exact posterior $Q^i(Z)=P_{\theta^{(t)}}(Z|X^i)$ for each sample $X^i$ in the \textbf{E}-step, so we may resort to approximations of the true posterior. In particular, we generate approximate sample paths from $P_\theta^{(t)}(Z|X^i)$ using Metropolis-Hastings type MCMC algorithm. 

In the \textbf{M}-step, finding the argmax is difficult since the expectation over the approximate posterior does not have a closed-form expression. Instead, we perform a few gradient steps with respect to $\theta$, which requires Monte-Carlo estimates of the gradients of right-hand size of Eq.~\eqref{eq:elbo_2}. The next section explains how to compute these gradient estimates. 

\subsection{Gradient Computations}
Starting with Eq.~\eqref{eq:elbo_2}, observe that 
$$\partial_\theta\,\text{ELBO}=-\partial_\theta\E[Q(Z)]{\log\frac{dQ(Z)}{dP_\theta(Z)}}=-\partial_\theta\E[Q(Z)]{\log\frac{dP^*(Z)}{dP_\theta(Z)}},$$
where $P^*(Z)$ is the path measure induced by the SDE
$$dZ_t=\sigma(Z_t,t)\,dB_t.$$
This identity holds because, during the M-step, the variational distribution $Q(Z)$ is fixed and does not depend on $\theta$. Furthermore, the data likelihood term $\log P(X|Z)$ is also independent of $\theta$. 

Assuming regularity conditions (as before), we may write
\begin{align*}
\partial_\theta\,\text{ELBO}&=-\partial_\theta\E[Q(Z)]{\int_0^T\frac{b_\theta}{\sigma^2}\,dZ_t-\frac{1}{2}\int_0^T\frac{b_\theta^2}{\sigma^2}\,dt}\\&=\E[Q(Z)]{\int_0^T\frac{b_\theta\,\partial_\theta b_\theta}{\sigma^2}\,dt-\int_0^T\frac{\partial_\theta b_\theta}{\sigma^2}\,dZ_t}.
\end{align*}
To obtain Monte Carlo estimates of this gradient, we substitute discretized increments $dZ_{t_i}\approx(Z_{t_{i+1}}-Z_{t_i})$ and evaluate the resulting expression w.r.t. sample paths $Z\sim Q(Z)$ generated in the E-step. Averaging over these trajectories and over batched samples yields an unbiased estimator of the gradient (up to discretization error).

\section{Variational Inference Gradient Computations}\label{sec:variational_inference}
In Section \ref{sec:gradient_computations}, we give the high-level ideas of how gradients of the ELBO are computed. Here, we dive into the details of those gradient computations.

\subsection{ELBO derivation (Proof of Theorem~\ref{thm:elbo})}\label{sec:vi_elbo_derivation}
We begin by deriving the Evidence lower bound (ELBO) for the variational formulation as shown in Eq.~\eqref{eq:elbo_vi}. Starting from the general ELBO expression in Eq.~\eqref{eq:elbo}, we replace the variational measure $Q(Z)$ with the conditional path measure $Q_\phi(Z|X)$ that is induced by the posterior SDE in Eq.~\eqref{eq:posterior_sde}. Recall the version of the Girsanov's theorem adapted to our notations:
\begin{theorem}[Girsanov's theorem]
Let $(\Omega,\F,(\F_t)_{t\in[0,T]},\mathbb{P})$be a filtered probability space for the standard Brownian motion $B$. Consider the prior intensity diffusion
$$dZ_t=b_\theta(Z_t,t)\,dt +\sigma(Z_t,t)\,dB_t,\quad Z_0=z_0,$$

where $\sigma(\cdot,\cdot)\in\mathbb{R}$ and assume the usual conditions ensuring a unique strong solution. 

Let $u$ be an $\mathbb{R}$-valued process such that the exponential local martingale 
$$M_t:=\exp\left(\int_0^tu_s\,dB_s-\frac{1}{2}\int_0^t\lVert u_s\rVert^2\,ds\right)$$
is a true martingale on $[0,T]$ (e.g. by Novikov's condition). Define a new measure $\mathbb{Q}$ by 
$$\frac{d\mathbb{Q}}{d\mathbb{P}}\Bigg|_{\F_T}=M_T.$$
Then the process
$$\tilde{B}:=B_t-\int_0^tu_s\,ds$$
is a $\mathbb{Q}$-Brownian motion, and any $(\F_t)$-adapted process $Z$ satisfying 
$$dZ_t=\left(b_\theta(Z_t,t)+\sigma(Z_t,t)u_t\right)dt+\sigma(Z_t,t)\,dB_t$$
under $\mathbb{P}$ also satisfies
$$dZ_t=b_\theta(Z_t,t)\,dt+\sigma(Z_t,t)\,d\tilde{B}_t$$
under $\mathbb{Q}$.

Moreover,
$$\KL{\mathbb{Q}||\mathbb{P}}=\E[\mathbb{Q}]{\int_0^Tu_s\,dB_s-\frac{1}{2}\int_0^T\lVert u_s\rVert^2\,ds}.$$
\end{theorem}

By the theorem, the resulting KL divergence term admits a closed-form expression:
\begin{align*}
\KL{Q_\phi(Z|X)||P_\theta(Z)}&=\E[Q_\phi(Z|X)]{\log\left(\frac{dQ_\phi(Z|X)}{dP_\theta(Z)}\right)}\\
&=\E[Q_\phi(Z|X)]{\int_0^{T'}u_\beta(Z_t,t,T',X)\,dB_t-\frac{1}{2}\int_0^{T'}u_\beta^2(Z_t,t,T',X)\,dt}.
\end{align*}

Let $\mathbb{Q}$ denote the probability measure corresponding to $Q_\phi(Z|X)$. Under $\mathbb{Q}$, the process
$$\tilde{B}_t=B_t-\int_0^t u_\beta(Z_t,t,T',X)\,dt$$
is a standard Brownian motion. Therefore, we may rewrite the stochastic integral as
$$\int_0^{T'}u_\beta(Z_t,t,T',X)\,dB_t=\int_0^{T'}u_\beta(Z_t,t,T',X)\,d\tilde{B}_t+\int_0^{T'}u_\beta^2(Z_t,t,T',X)\,dt.$$
Substituting this expression back into the KL divergence and using the martingale property of $\tilde{B}_t$ under $\mathbb{Q}$, we obtain
$$\KL{Q_\phi(Z|X)||P_\theta(Z)}=\E[Q_\phi(Z|X)]{\frac{1}{2}\int_0^{T'}u_\beta^2(Z_t,t,T',X)\,dt},$$
which directly yields the ELBO expression in Eq.~\eqref{eq:elbo_vi}. 

\subsection{Gradient computations}\label{sec:vi_gradients}
We now turn to the computation of gradients of the ELBO with respect to the model parameters $\theta$ and $\beta$. Under suitable regularity conditions assumptions, we may interchange differentiation and expectation (e.g., by Fubini's theorem), leading to the following pathwise gradient representations:
\begin{align*}
\partial_\theta\,\text{ELBO}&=\E{\sum_{0<\tau_i\le T'}\frac{\partial_\theta Z_{\tau_i}}{Z_{\tau_i}}-\int_0^{T'}\partial_\theta Z_t\,dt-\int_0^{T'}u_\beta\frac{\partial u_\beta}{\partial Z_t}\partial_\theta Z_t\,dt},\\
\partial_\beta\,\text{ELBO}&=\E{\sum_{0<\tau_i\le T'}\frac{\partial_\beta Z_{\tau_i}}{Z_{\tau_i}}-\int_0^{T'}\partial_\beta Z_t\,dt-\int_0^{T'}u_\beta\left(\frac{\partial u_\beta}{\partial Z_t}\partial_\beta Z_t+\partial_\beta u_\beta\right)\,dt}.
\end{align*}

The expectations above are taken with respect to the joint law of $Z_t$ and its parameter sensitivities. These evolve according to the following set of SDEs. $Z_t$ satisfies Eq.~\eqref{eq:posterior_sde}, while the sensitivities with respect to $\theta$ and $\beta$ satisfy
\begin{align} 
\label{eq:jac-theta}
d\partial_\theta Z_t&=\left[\partial_\theta b_\theta+\frac{\partial b_\theta}{\partial Z_t}\partial_\theta Z_t+\sigma\frac{\partial u_\beta}{\partial Z_t}\partial_\theta Z_t+\frac{\partial\sigma}{\partial Z_t}u_\beta\partial_\theta Z_t\right]dt+\frac{\partial\sigma}{\partial Z_t}\partial_\theta Z_t\,dB_t,\, \partial_\theta Z_t=\mathbf{0},\\
\label{eq:jac-beta}
d\partial_\beta Z_t&=\left[\frac{\partial b_\theta}{\partial Z_t}\partial_\beta Z_t+\sigma\frac{\partial u_\beta}{\partial Z_t}\partial_\beta Z_t+\sigma\partial_\beta u_\beta+\frac{\partial\sigma}{\partial Z_t}u_\beta\partial_\beta Z_t\right]dt+\frac{\partial\sigma}{\partial Z_t}\partial_\beta Z_t\,dB_t,\, \partial_\beta Z_t=\mathbf{0}.
\end{align}
To obtain Monte Carlo estimates of the gradients, we sample $m$ i.i.d. discretized Brownian paths. For each path, simulate $Z_t$, $\partial_\theta Z_t$, and $\partial_\beta Z_t$ using a shared Brownian path $B_t$ with Euler-Maruyama discretization. Substituting these simulated trajectories into the expressions above yields unbiased estimators of the gradients, other than the discretization bias introduced by the Euler scheme.

\section{About the Bank Dataset}\label{sec:call_data}
The dataset~\citep{seelab2024data} originates from an undisclosed U.S. bank. With up to 300,000 calls per day, it provides a rich collection of information, including call time, call duration, and various categorical and continuous data. For our analysis, we focus solely on the initial arrival times of the calls. Accordingly, the dataset records the number of calls occurring within each minute of the day. Given the large volume of calls, we first evenly distribute calls within each minute and then apply independent thinning to each event with probability $p=0.001$.

\subsection{Inference with Overdispersed Data}~\label{sec:call_data_expt}

\begin{figure}[t]
    \centering
    \begin{subfigure}[b]{0.48\textwidth}
        \centering
        \includegraphics[width=\textwidth]{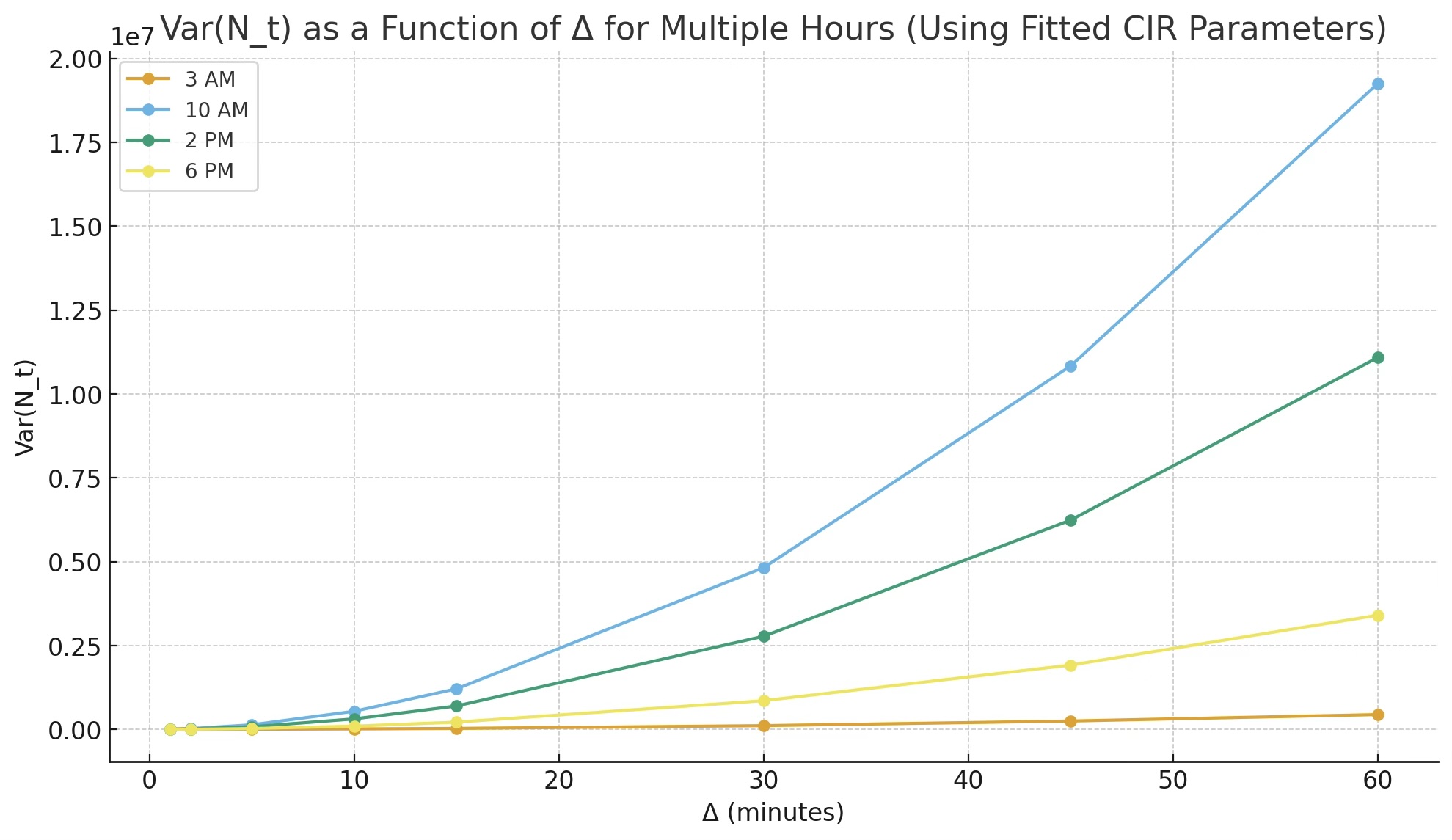}
        \caption{Variance of event counts as a function of $\Delta$, with CIR intensity model.}\label{fig:overdisperse-counts}
    \end{subfigure}~
    \begin{subfigure}[b]{0.48\textwidth}
        \centering
        \includegraphics[width=\textwidth]{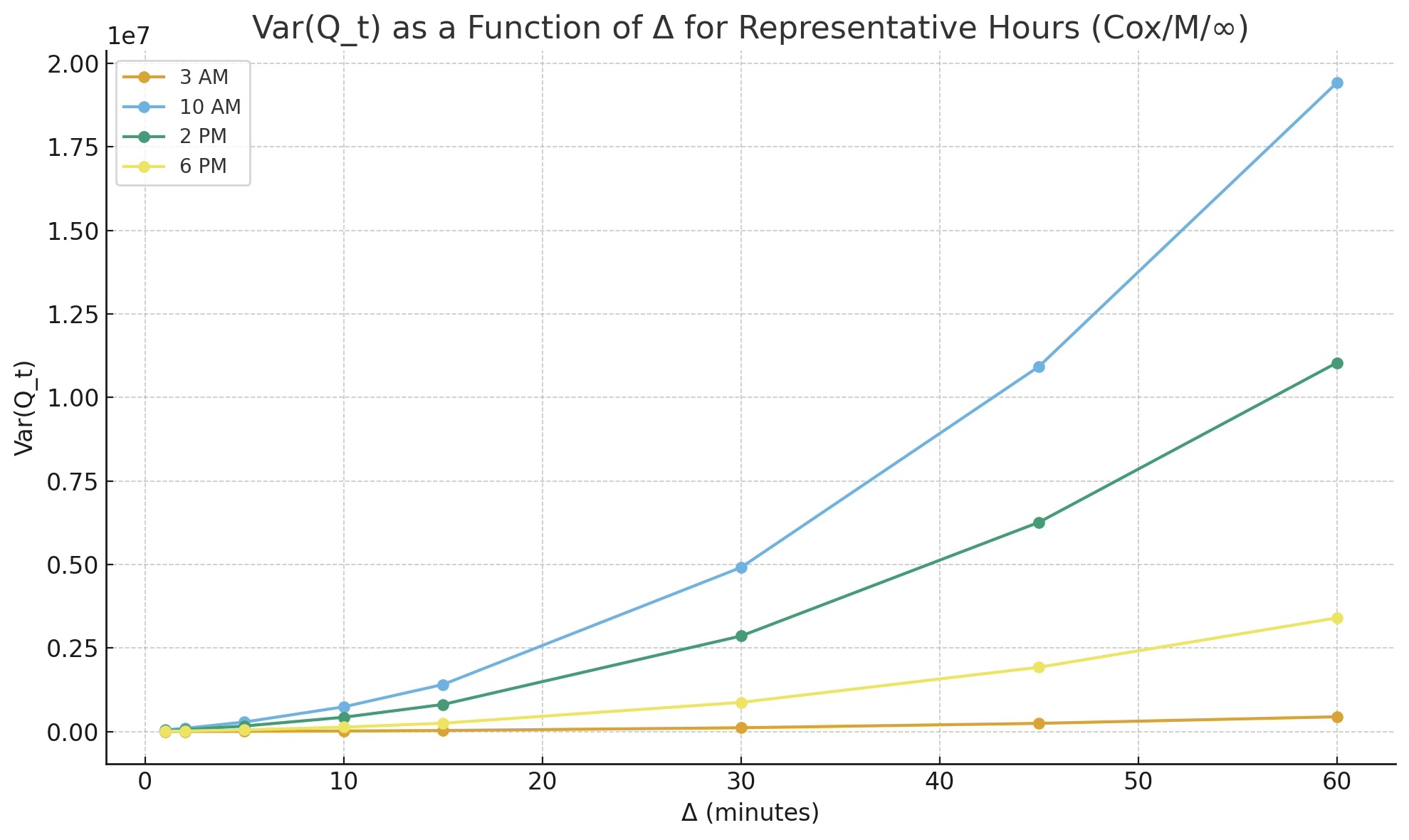}
        \caption{Variance of Cox$/M/\infty$ queue length as a function of $\Delta$.}~\label{fig:overdisperse-queue}
    \end{subfigure}
    \caption{Overdispersion affects downstream inferences.}
    \label{fig:overdisperse}
\end{figure}
We now describe a simple experiment demonstrating overdispersion in the U.S. Bank dataset. As Figure~\ref{fig:placeholder} shows, the estimated mean and variance of arrival counts over 10-minute intervals differ by more than an order of magnitude. The ratio of the variance to the mean is called the index of dispersion~\citep{cox1966statistical}. Under a nonhomogeneous Poisson process, the index of dispersion would be uniformly equal to 1 by definition, so the discrepancy observed in the dataset is direct evidence that the data-generating process is not a simple Poisson point process.

Overdispersion has immediate implications for downstream inference. For a nonhomogeneous Poisson process with deterministic intensity $(\lambda_t)$, the variance of the count $N(t) := \sum_{i \geq 1} \mathbbm{1}_{\tau_i \leq t}$ over an interval $[t, t+\Delta)$ satisfies
\begin{align*}
    \text{Var}_{\mu}(N(t+\Delta) - N(t)) = \mathbb{E}_{\mu}[N(t+\Delta) - N(t)] \approx \lambda(t)\Delta.
\end{align*}
That is, variance grows linearly in $\Delta$ at the same rate as the mean. For a Cox process with stochastic intensity $(Z_t)$ satisfying $\mathbb{E}[Z(t)] = \lambda(t)$, the law of total variance gives
\begin{align*}
    \text{Var}_{\mu}(N(t+\Delta) - N(t)) 
    &= \mathbb{E}_{Z}\left[\text{Var}_{N}\left(N(t+\Delta) - N(t) \mid Z\right)\right]\\ &\qquad\qquad 
    + \text{Var}_{Z}\left(\mathbb{E}_{N}\left[N(t+\Delta) - N(t) \mid Z\right]\right) \\
    &\approx \lambda(t)\Delta + \Delta^2\,\text{Var}_{Z}(Z(t)).
\end{align*}
The additional term $\Delta^2\,\text{Var}_Z(Z(t))$, absent in the Poisson case, reflects the randomness of the latent intensity and causes the variance to grow quadratically in $\Delta$.

Figure~\ref{fig:overdisperse-counts} plots the empirical variance of arrival counts from the U.S. Bank dataset as a function of $\Delta$, at four representative hours: 3AM, 10AM, 2PM, and 6PM. The variance grows  nonlinearly, consistent with the Cox process prediction and inconsistent with any Poisson model regardless of how its intensity is specified. We also fit a CIR diffusion intensity model to each time period. Figure~\ref{fig:overdisperse-queue} then plots the variance of the queue length in a $\text{Cox}/M/\infty$ queue driven by the fitted Cox process. The nonlinear growth persists in the queue length as well, illustrating that overdispersion propagates into downstream inference: a model that does not explicitly represent the stochastic intensity will systematically misrepresent the variability of quantities that depend on it.

\section{Architecture}
\begin{figure}[b]
  \centering
  \includegraphics[width=\textwidth]{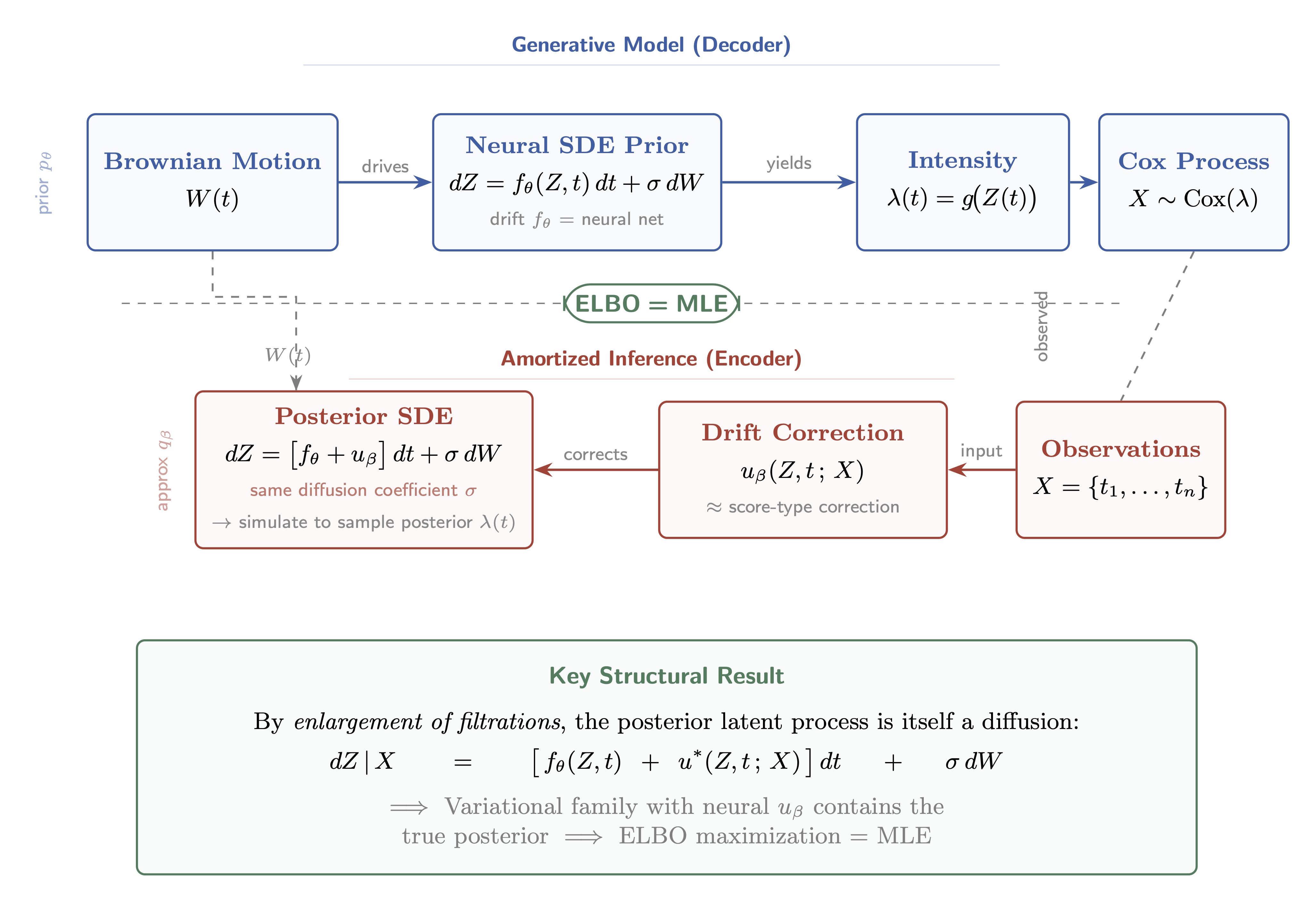}
  \caption{Amortized variational inference architecture for Neural Diffusion Intensity Models.}
  \label{fig:framework}
\end{figure}
\end{document}